\documentclass[11pt]{article}

\usepackage[preprint]{acl}
\usepackage{times}
\usepackage{latexsym}
\usepackage[T1]{fontenc}
\usepackage[utf8]{inputenc}
\usepackage{microtype}
\usepackage{inconsolata}
\usepackage{graphicx}
\usepackage{multirow}
\usepackage{subcaption}
\usepackage{placeins}
\usepackage{tcolorbox}
\usepackage{listings}
\usepackage{xcolor}

\title{Mechanistic Interpretability of Large-Scale Counting in LLMs\\ through a System-2 Strategy}

\author{\textbf{Hosein Hasani},
  \textbf{Mohammadali Banayeeanzade},
  \textbf{Ali Nafisi},
  \textbf{Sadegh Mohammadian},\\
  \textbf{Fatemeh Askari},
  \textbf{Mobin Bagherian},
  \textbf{Amirmohammad Izadi},\\
  \textbf{and Mahdieh Soleymani Baghshah}\\
  \vspace{9mm} 
  Sharif University of Technology
}

\begin{document}
\maketitle
\begin{abstract}

Large language models (LLMs), despite strong performance on complex mathematical problems, exhibit systematic limitations in counting tasks. This issue arises from the architectural limits of transformers, where counting is performed across layers, leading to degraded precision for larger counting problems due to depth constraints. To address this limitation, we propose a simple test-time strategy inspired by System-2 cognitive processes that decomposes large counting tasks into smaller, independent sub-problems that the model can reliably solve. We evaluate this approach using observational and causal mediation analyses to understand the underlying mechanism of this System-2-like strategy. Our mechanistic analysis identifies key components: latent counts are computed and stored in the final item representations of each part, transferred to intermediate steps via dedicated attention heads, and aggregated in the final stage to produce the total count. Experimental results demonstrate that this strategy enables LLMs to surpass architectural limitations and achieve higher accuracy on large-scale counting tasks. This work provides mechanistic insight into System-2 counting in LLMs and presents a generalizable approach for improving and understanding their reasoning behavior.

\end{abstract}

\section{Introduction}

Counting is a fundamental cognitive operation that underpins a wide range of reasoning tasks, from basic arithmetic to more complex forms of quantitative analysis~\citep{feigenson2004core,dehaene2011numbersense}. In large language models (LLMs), the ability to count is important for controlled generation such as length-constrained summarization, sequential enumeration, and broader numerical/arithmetic reasoning \citep{retkowski-waibel-2025-zero,hou2025sequentialenumeration,yang2024numbercookbook,gambardella-etal-2024-language}.

Recent research has provided insights into the counting mechanism of LLMs, demonstrating that these models use a layerwise internal counting process where numerical information is progressively accumulated across transformer layers~\citep{hasani2025countscope}. However, due to the depth limits, this progressive counting mechanism becomes saturated as the number of items increases. Complementary work on numerical representations suggests that LLMs encode numbers in a compressed, sublinear manner, similar to the human logarithmic mental number line~\citep{alquboj2025numberrepresentations,dehaene2011numbersense}. While numerical order is preserved, representational resolution decreases with magnitude, explaining the reduced precision for large values.

\begin{figure}[t]
    \centering
    \includegraphics[width=0.99\linewidth]{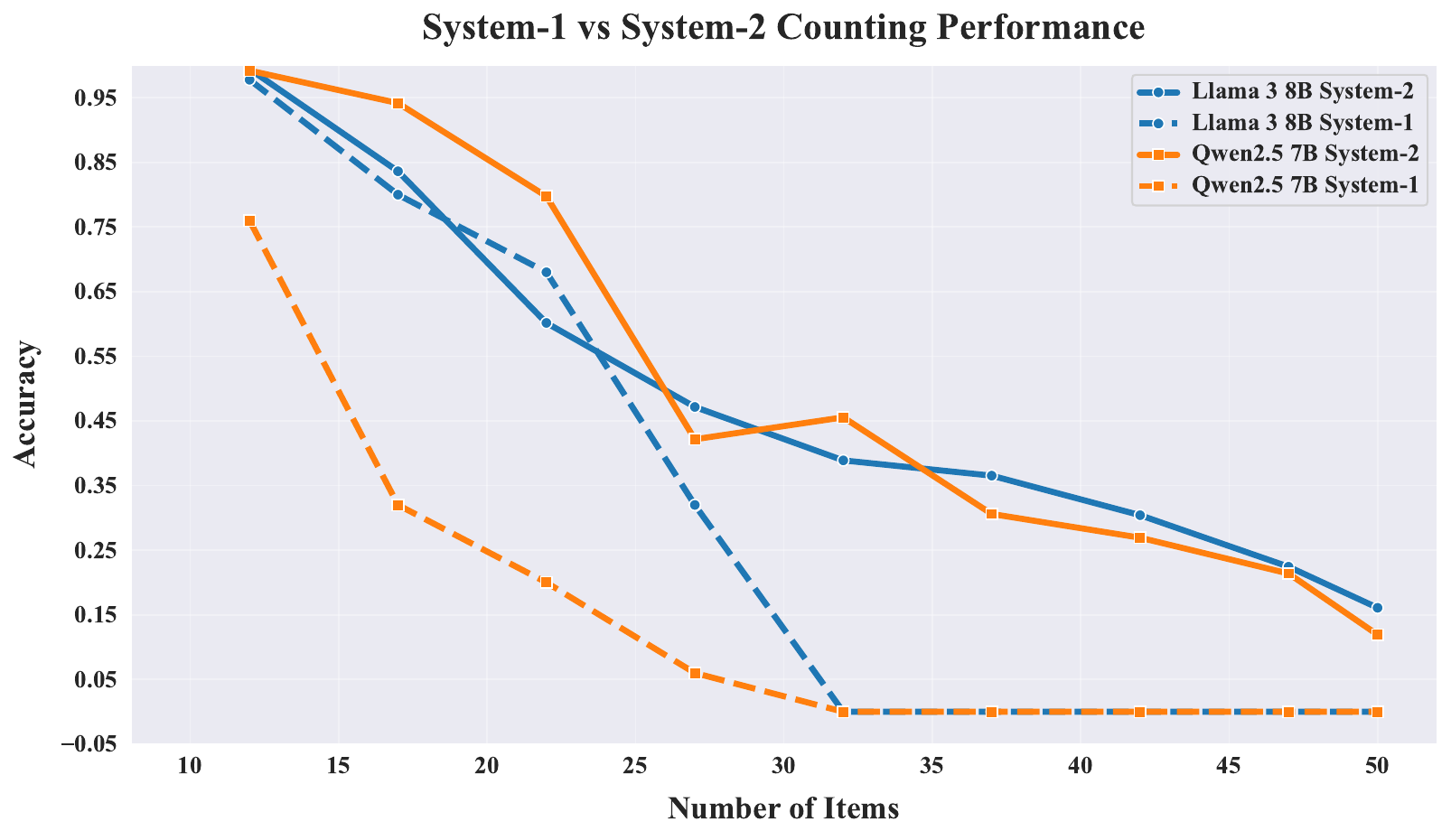}
    \vspace{-4pt}
    \caption{
    System-1 vs. System-2 counting performance as a function of problem size.
    System-1 performance degrades rapidly and collapses beyond approximately 30 items, reflecting the bounded capacity of the model’s internal counter.
    In contrast, System-2 counting maintains high accuracy across the entire range by decomposing the task into small solvable sub-problems and aggregating the results.
    }
    \vspace{-2.0mm}
    \label{fig:system2_vs_system1}
\end{figure}

\begin{figure*}[t]
    \centering
    \includegraphics[width=0.95\linewidth]{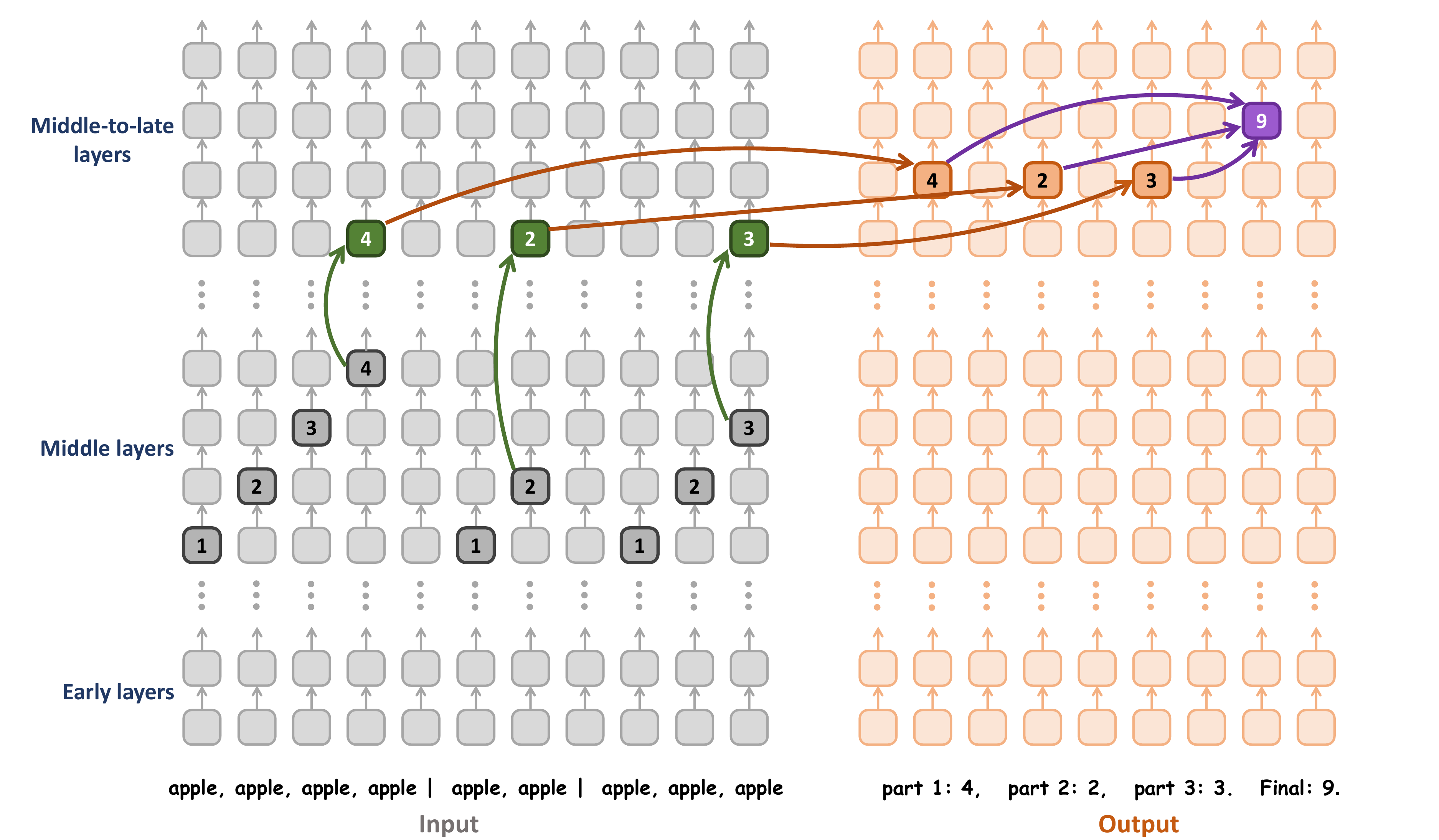}
    \vspace{-4pt}
    \caption{
    \textbf{Internal mechanism of System-2 test-time counting in LLMs.}
    A large counting task is divided into smaller partitions using an external separator (|). Within each partition, the model performs implicit System-1 counting, where count information accumulates token-by-token and is localized at the final item or separator token (gray blocks). The final count information is transferred via residual streams (green arrows) and stored in the middle-to-late layers (green blocks). These partition-level counts (e.g., 4, 2, 3) are then transferred (orange arrows) through attention pathways to explicit reasoning tokens that report intermediate results (orange blocks). Finally, the intermediate counts are aggregated (purple arrows) to produce the final answer. By keeping each sub-task within the model’s reliable counting range, this System-2 procedure removes the upper bound imposed by the model’s architectural limitations.
    }
    \vspace{-2.0mm}
    \label{fig:system2_counting}
\end{figure*}

These findings suggest that LLMs struggle to accurately count large numbers of items, with performance typically degrading for two- and three-digit counts~\citep{zhang2024countingabilitytokenization,yehudai2024whencantransformerscount,fu2024lettercounting}. This limitation reflects a fundamental architectural constraint rather than a lack of training data or supervision, as more layers are required to count a larger number of items~\citep{vaswani2017attention,yehudai2024whencantransformerscount,golkar2024contextualcounting}. We argue that this failure stems from the model's reliance on a System-1–like processing approach, which is fast, automatic, and capacity-limited~\citep{kahneman2011thinking,dehaene2011numbersense}.

To address this issue, we propose a simple test-time strategy that adopts a System-2--like approach to counting (Figure~\ref{fig:system2_vs_system1})~\citep{kahneman2011thinking}. Instead of relying on the model's internal counting mechanisms, which are constrained by architectural limits, our approach decomposes large counting tasks into smaller, independent sub-tasks~\citep{radhakrishnan2023questiondecomposition,qharabagh2024lvlmcount}. Each sub-task, containing a manageable number of items, can be reliably counted by the model. The results from each sub-task are then aggregated to produce the final count. This approach mirrors human cognitive strategies, where complex problems are broken down into simpler, easier-to-solve sub-problems~\citep{dehaene2011numbersense}. In this study, we use System-1 and System-2 as high-level operational abstractions inspired by human cognition~\citep{kahneman2011thinking}. Specifically, we refer to System-1 as the model's single-pass, layer-wise processing, and to System-2 as a more deliberate procedure that extends computation across token generation beyond a fixed layer-wise computation.

Our behavioral experiments on various LLMs demonstrate the effectiveness of this strategy in overcoming architectural limitations without requiring model modifications or fine-tuning. Additionally, we provide a detailed mechanistic interpretation of how it functions within LLMs. Using attention analysis and causal mediation techniques~\citep{heimersheim2024activationpatching,geiger2021causalabstractions,ghandeharioun2024patchscopes,zhang2024activationpatching_bestpractices}, we trace the flow of numerical information across the model and identify the mechanisms mediating the System-2 counting process.
Figure~\ref{fig:system2_counting} provides an overview of the main components of the System-2 counting mechanism and its internal information flow.
This work offers a new perspective on both improving and understanding LLMs' reasoning capabilities, with a focus on a fundamental cognitive task.

\section{Problem Setup and Methodology}

We follow the standard counting framework used in prior research~\citep{hasani2025countscope}. In this setup, a list of repeated items, such as fruits or animals, is presented to the model, and the task is to report the total number of items. For example, given a context like ``apple, apple, apple, \dots'', the model must output the total number of items in the list. Previous work has shown that LLMs exhibit high accuracy for small counts (fewer than 10 items), but that performance deteriorates as the number of items increases beyond this range~\citep{zhang2024countingabilitytokenization,fu2024lettercounting}. This suggests that the model’s counting ability is limited by the depth of the transformer architecture and its internal counting mechanism~\citep{yehudai2024whencantransformerscount}.

To overcome this limitation, motivated by prior work on partitioning images in visual reasoning tasks~\citep{izadi2025viser}, we introduce a simple strategy that explicitly partitions the input list into smaller sub-problems. We use an external separator (|) to divide the list into smaller partitions. For instance, a structured context with three partitions is:
``apple, apple, apple, apple, apple | apple, apple, apple | apple, apple, apple, apple, apple, apple''.

The model is then instructed to first count the items in each partition and to aggregate the partial counts to produce the final result. This ensures that each sub-problem remains within the model’s reliable counting regime. The number of items in each partition is chosen randomly from a range that the model can handle accurately. Models with deeper architectures can reliably process larger partitions.
This strategy is based on test-time scaling by leveraging the LLM’s inherent capabilities and does not require fine-tuning or any external tools.

Based on the input structure and output format, we consider four baselines in our study. Two input formats are used: an unstructured context with comma-separated items and a structured context with partitions separated by vertical bar symbol (|). For generation, we evaluate two output variants: a short-answer format, where only the final count is produced (corresponding to an immediate System-1-like process), and a Chain-of-Thought (CoT)~\citep{wei2022cot} format, where intermediate reasoning steps are included before the final answer (corresponding to a System-2-like process).
Full details of the input formats, partitioning, and prompting strategies are provided in Appendix~\ref{sec:behavioral_setup}.

\section{Behavioral Results}
\label{sec:behavioral}

\begin{figure}[t]
    \centering
    \includegraphics[width=0.999\linewidth]{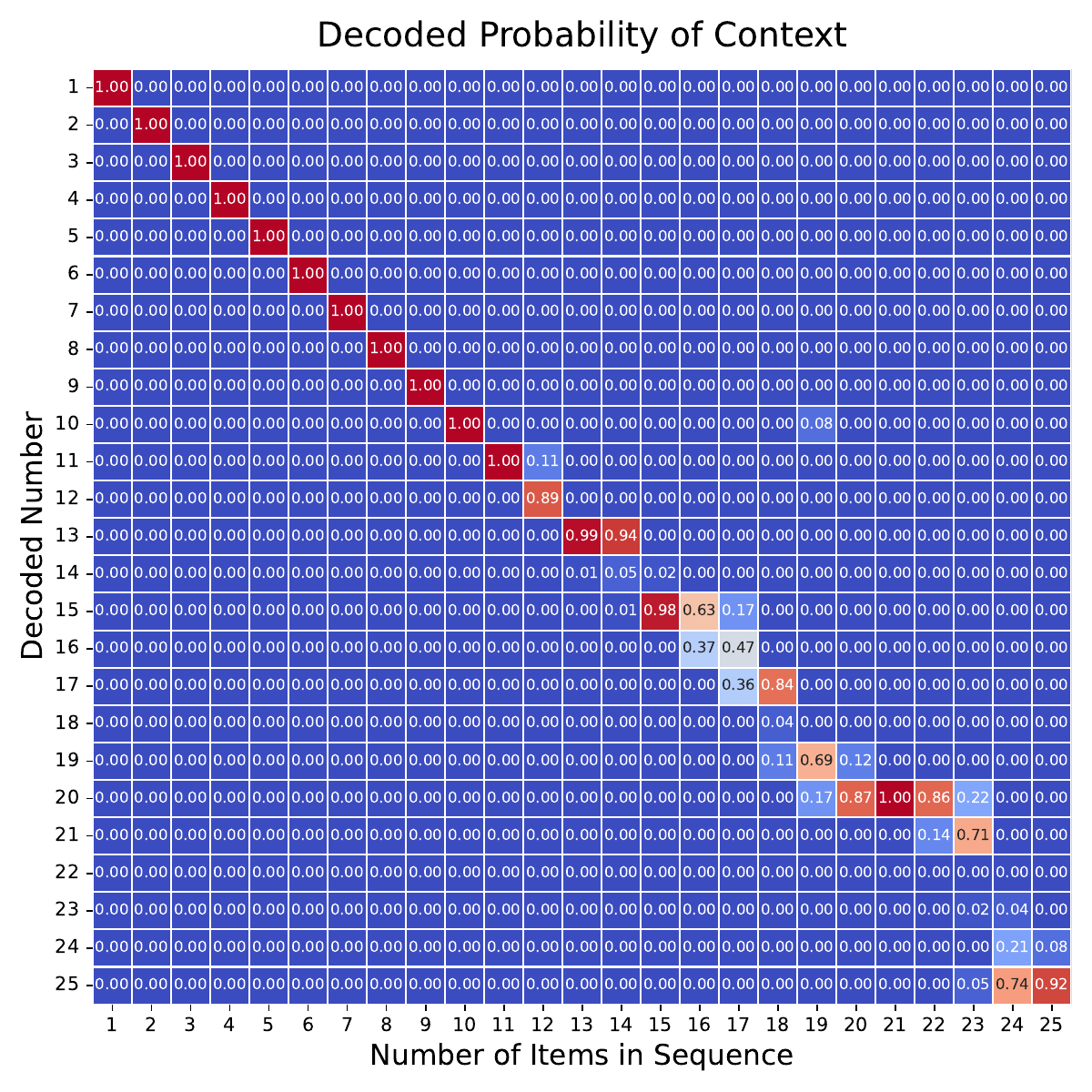}
    \vspace{-4pt}
    \caption{
    Decoded output probabilities for the unstructured baseline method on Qwen2.5-7B.
    The heatmap shows the decoded probabilities of model outputs, averaged over different item types, for target counts ranging from 1 to 25. As the count increases beyond 10, the diagonal entries gradually fade, indicating reduced model confidence.
    }
    \vspace{-2.0mm}
    \label{fig:output_probs:baseline}
\end{figure}

To illustrate how LLMs lose counting precision in long contexts, we first measure the average prediction probability of Qwen2.5-7B~\citep{qwen25} across different context sizes. As shown in Figure~\ref{fig:output_probs:baseline}, model confidence decreases as the number of items exceeds 10. We also observe systematic biases toward more frequent numbers. For example, target counts of 16 and 24 are often predicted as 15 or 25, with 21 and 22 typically predicted as 20.

\begin{table*}[htb]
\centering
\resizebox{0.82\linewidth}{!}{
\begin{tabular}{lcc|cccc|cccc}
\hline
\textbf{Model} & \textbf{Input} & \textbf{Output} 
& \multicolumn{4}{c|}{\textbf{Accuracy}} 
& \multicolumn{4}{c}{\textbf{MAE}} \\ 
& & 
& \textbf{11-20} & \textbf{21-30} & \textbf{31-40} & \textbf{41-50}
& \textbf{11-20} & \textbf{21-30} & \textbf{31-40} & \textbf{41-50} \\ \hline

\multirow{4}{*}{\textbf{Qwen2.5-7B}}
& \multirow{2}{*}{Unstructured} & w/o steps
& 0.38 & 0.13 & 0.06 & 0.00
& 0.88 & 2.19 & 5.29 & 10.50 \\
&  & w/ steps
& 0.45 & 0.11 & 0.03 & 0.00
& 0.72 & 2.44 & 5.97 & 9.68 \\
\cline{2-11}
& \multirow{2}{*}{Structured} & w/o steps
& 0.20 & 0.13 & 0.05 & 0.01
& 3.56 & 3.54 & 4.33 & 6.35 \\
&  & w/ steps
& \textbf{0.95} & \textbf{0.61} & \textbf{0.38} & \textbf{0.24}
& \textbf{0.07} & \textbf{1.36} & \textbf{1.53} & \textbf{2.18} \\ \hline

\multirow{4}{*}{\textbf{Llama3-8B}}
& \multirow{2}{*}{Unstructured} & w/o steps
& 0.80 & 0.49 & 0.02 & 0.00
& \textbf{0.20} & \textbf{0.65} & 4.93 & 11.92 \\
&  & w/ steps
& 0.29 & 0.34 & 0.00 & 0.00
& 0.74 & 0.82 & 5.00 & 11.44 \\
\cline{2-11}
& \multirow{2}{*}{Structured} & w/o steps
& 0.08 & 0.05 & 0.03 & 0.01
& 6.62 & 5.75 & 5.96 & 6.62 \\
&  & w/ steps
& \textbf{0.84} & \textbf{0.54} & \textbf{0.38} & \textbf{0.26}
& 0.30 & 0.70 & \textbf{1.21} & \textbf{2.20} \\ \hline

\multirow{4}{*}{\textbf{Gemma3-27B}}
& \multirow{2}{*}{Unstructured} & w/o steps
& \textbf{1.00} & 0.70 & 0.40 & 0.00
& \textbf{0.00} & \textbf{0.30} & \textbf{1.00} & 4.90 \\
&  & w/ steps
& \textbf{1.00} & 0.50 & 0.30 & 0.00
& \textbf{0.00} & 0.60 & 1.40 & 4.70 \\
\cline{2-11}
& \multirow{2}{*}{Structured} & w/o steps
& 0.30 & 0.05 & 0.00 & 0.17
& 2.10 & 10.95 & 7.33 & 12.67 \\
&  & w/ steps
& \textbf{1.00} & \textbf{0.85} & \textbf{0.55} & \textbf{0.50}
& \textbf{0.00} & 0.35 & 2.15 & \textbf{2.25} \\ \hline

\end{tabular}
}
\caption{Average accuracy and mean absolute error (MAE) across \textbf{open-source models} for context sizes from 11 to 50. Each model is evaluated on structured and unstructured inputs, with and without intermediate reasoning steps. MAE complements exact-match accuracy by measuring how far incorrect predictions are from the true count.}
\label{tab:behavioral:open_source}
\end{table*}

\begin{table*}[htb]
\centering
\resizebox{0.97\linewidth}{!}{
\begin{tabular}{lcc|ccccc|ccccc}
\hline
\textbf{Model} & \textbf{Input} & \textbf{Output} 
& \multicolumn{5}{c|}{\textbf{Accuracy}} 
& \multicolumn{5}{c}{\textbf{MAE}} \\ 
& & 
& \textbf{51-60} & \textbf{61-70} & \textbf{71-80} & \textbf{81-90} & \textbf{91-100} 
& \textbf{51-60} & \textbf{61-70} & \textbf{71-80} & \textbf{81-90} & \textbf{91-100} \\ \hline

\multirow{4}{*}{\textbf{GPT-4o}}
& \multirow{2}{*}{Unstructured} & w/o steps 
& 0.70 & 0.54 & 0.56 & 0.18 & 0.24
& 0.36 & 1.42 & 0.72 & 3.62 & 4.26 \\
&  & w/ steps  
& 0.58 & 0.53 & 0.40 & 0.16 & 0.26 
& 1.10 & 1.78 & 1.72 & 3.20 & 3.64 \\
\cline{2-13}
& \multirow{2}{*}{Structured} & w/o steps 
& 0.37 & 0.31 & 0.10 & 0.11 & 0.11
& 1.04 & 1.53 & 3.22 & 2.64 & 3.03 \\
&  & w/ steps  
& \textbf{0.96} & \textbf{0.91} & \textbf{0.87} & \textbf{0.83} & \textbf{0.86}
& \textbf{0.04} & \textbf{0.10} & \textbf{0.16} & \textbf{0.22} & \textbf{0.18} \\ \hline

\multirow{4}{*}{\textbf{Gemini-2.5-Pro}}
& \multirow{2}{*}{Unstructured} & w/o steps 
& 0.52 & 0.50 & 0.44 & 0.42 & 0.20
& 0.60 & 0.50 & 3.17 & 4.67 & 2.70 \\
&  & w/ steps  
& 0.95 & 0.80 & 0.72 & 0.60 & 0.60 
& \textbf{0.05} & 1.10 & 1.78 & 1.30 & 1.30 \\
\cline{2-13}
& \multirow{2}{*}{Structured} & w/o steps 
& 0.82 & 0.80 & 0.78 & 0.75 & 0.79
& 0.25 & 0.08 & 0.18 & 0.30 & 0.10 \\
&  & w/ steps  
& \textbf{0.97} & \textbf{0.95} & \textbf{0.95} & \textbf{0.91} & \textbf{0.91}
& 0.10 & \textbf{0.05} & \textbf{0.05} & \textbf{0.06} & \textbf{0.07} \\ \hline

\end{tabular}
}
\caption{Average accuracy and MAE across \textbf{closed-source models} for context sizes from 51 to 100. Each model is evaluated on structured and unstructured inputs, with and without intermediate reasoning steps.}
\label{tab:behavioral:closed_source}
\end{table*}

We then conduct more systematic experiments on both open-source and closed-source models, covering different model sizes, depths, training strategies, and tokenization schemes. For open-source models, we evaluate Qwen2.5-7B, Llama3-8B~\citep{llama3}, and Gemma3-27B~\citep{gemma3}, which have 28, 32, and 62 layers, respectively. In addition, we evaluate GPT-4o~\citep{Achiam2023GPT4} and Gemini-2.5-Pro~\citep{gemini2025pushing} as stronger proprietary models on longer contexts
\footnote{For these stronger closed-source models, performance is near-perfect for shorter contexts (below 50 items). We therefore use longer contexts, where the task is no longer saturated and differences between strategies become visible.}.
The corresponding results are reported in Tables~\ref{tab:behavioral:open_source} and~\ref{tab:behavioral:closed_source}. Overall, performance decreases as context size increases and is associated with model scale, especially depth, with larger and deeper models showing higher accuracy. For each model, we find that only one configuration consistently succeeds on long contexts: structured inputs combined with intermediate reasoning steps.

Surprisingly, encouraging CoT reasoning alone, without structured input, does not yield a notable improvement. Furthermore, structured input without intermediate steps is also ineffective and can be harmful in some cases. These results show that neither external structure nor reasoning alone is sufficient. Their combination is necessary to overcome large-scale counting failures. In many models, the unstructured short-answer baseline performs better than the structured short-answer baseline. This suggests that the model must first consolidate partial results in intermediate steps before aggregating them into the final answer through a two-stage process.

A likely explanation for the failure of the structured short-answer setting is related to the latent counting mechanism. Prior work showed that some models tend to output the maximum latent count observed in the context rather than the true total count~\citep{hasani2025countscope}. In our structured format, the counter resets after each partition, so the maximum latent count often matches the largest partition size instead of the total sum. Consistent with this view, for context sizes 11 to 30, 13\% of Qwen2.5-7B errors and 43.6\% of Llama3-8B errors exactly match the largest partition size.
In addition, the high MAE of this setting in Table~\ref{tab:behavioral:open_source} suggests systematic bias in erroneous predictions.
Our observational and causal analyses provide further insight into why models cannot simultaneously gather partial counts from the context and add them up without loss.

Finally, we analyze failure cases of the structured CoT setting by separating errors into intermediate partition counts and final aggregation. Table~\ref{tab:failure_analysis} shows that the main source of failure is the intermediate steps. This indicates that once correct partition counts are produced, most models can reliably add them.

\begin{table}[htb]
\centering
\resizebox{0.999\linewidth}{!}{
\begin{tabular}{lccc}
\hline
\textbf{Model} & \textbf{Total Acc} & \textbf{Final Step Acc} & \textbf{Interm. Acc} \\
\hline
Qwen2.5 7B     & 0.51 & 0.86 & 0.53 \\
Llama 3 8B     & 0.49 & 0.96 & 0.48 \\
Gemma 3 27B    & 0.71 & 0.93 & 0.76 \\
GPT-4o         & 0.89 & 1.00 & 0.89 \\
Gemini-2.5-Pro & 0.94 & 0.97 & 0.94 \\
\hline
\end{tabular}
}
\caption{Failure analysis of the structured CoT setting. Total accuracy measures exact final answers. Final-step accuracy measures correctness of the aggregation step conditioned on generated intermediate counts. Intermediate-step accuracy measures correctness of partition-level counts.}
\label{tab:failure_analysis}
\end{table}

\begin{figure*}[t]
    \centering
    \begin{subfigure}[t]{0.45\linewidth}
        \centering
        \includegraphics[width=\linewidth]{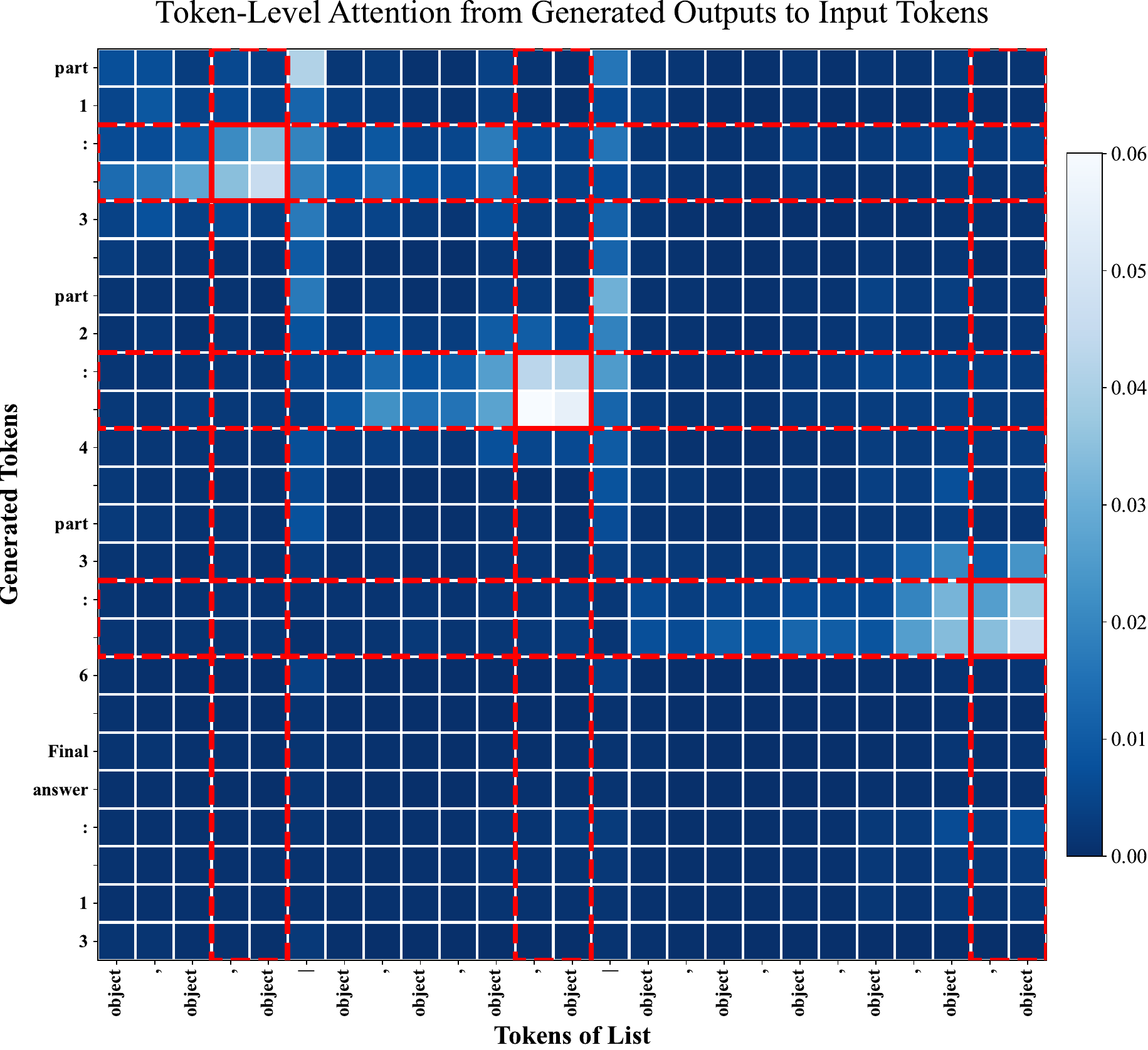}
        \caption{Attention of generated tokens to input tokens.}
        \label{fig:attn_reasoning}
    \end{subfigure}
    \hfill
    \begin{subfigure}[t]{0.44\linewidth}
        \centering
        \includegraphics[width=\linewidth]{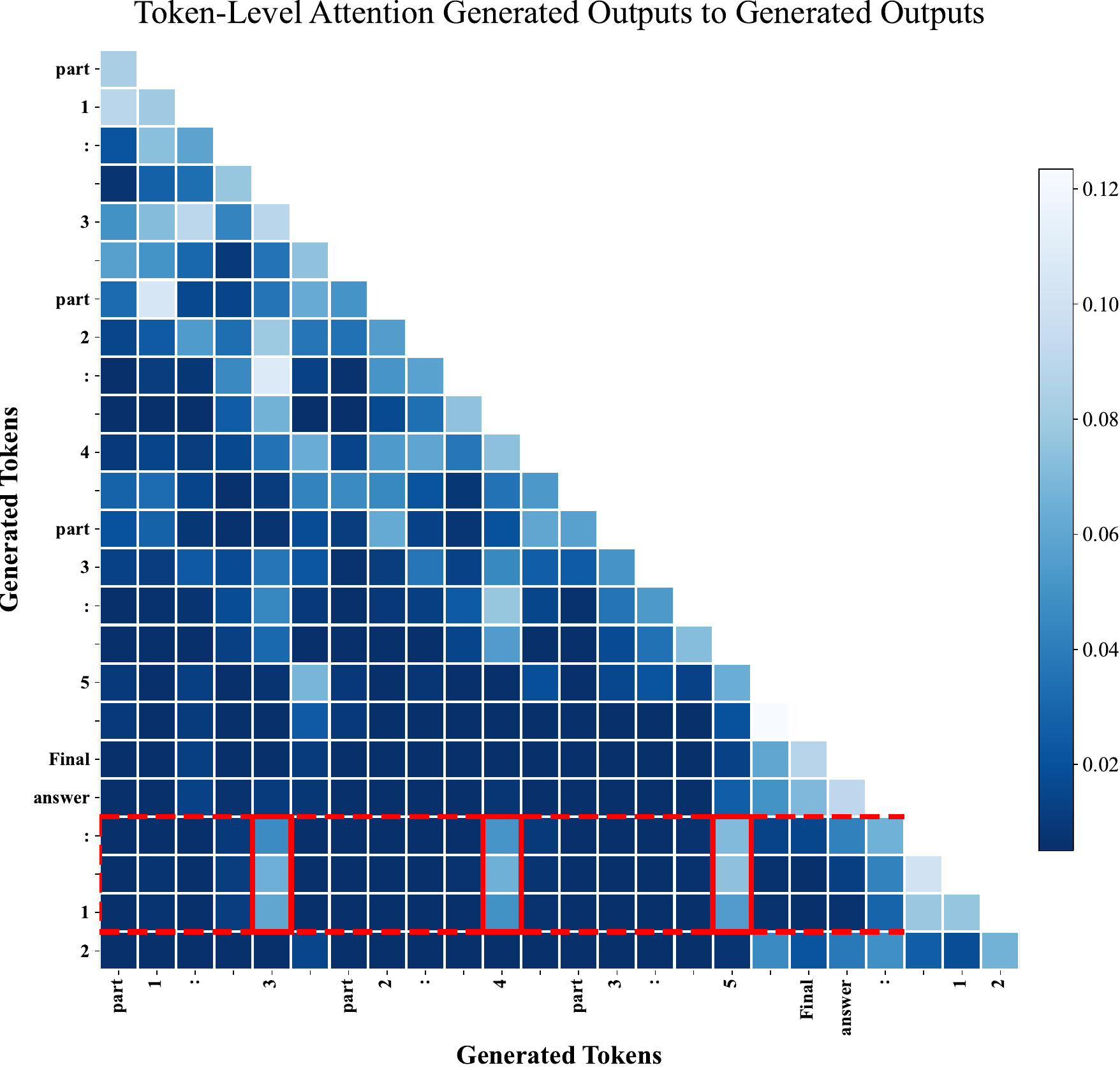}
        \caption{Attention of generated tokens to generated tokens.}
        \label{fig:attn_final}
    \end{subfigure}
    \vspace{-3pt}
    \caption{Attention patterns of selected tokens under the System-2 counting strategy. Attention values are averaged across layers 19 to 23, all heads, and all item types (e.g., different fruit names).}
    \label{fig:attention_system2}
\end{figure*}

Additional experiments on smaller and math-specialized models are presented in Appendix~\ref{sec:appendix_additional_models}, showing that the observed trends generalize across model scale and domain specialization. Robustness checks across tokenizers and alternative input structures are provided in Appendix~\ref{sec:robustness_input_structure}, and the overall trends hold across tokenizers and input formats, demonstrating that the System-2 mechanism is robust and generalizable.

\section{Attention Analysis}
\label{sec:attention_analysis}

\begin{figure}[t]
    \centering
    \includegraphics[width=0.99\linewidth]{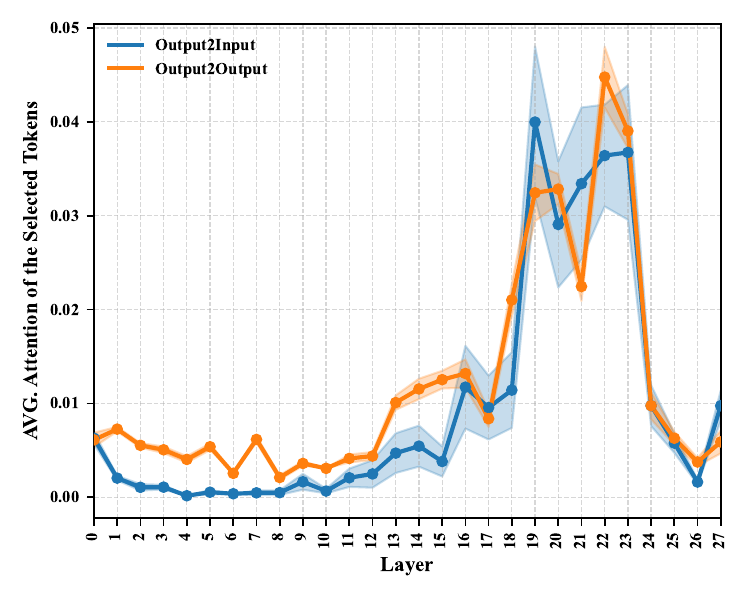}
    \vspace{-2pt}
    \caption{
    Average attention of correctly generated intermediate counts to the final item of their partition, alongside attention of the final answer to partition-level counts, across layers. Higher attention values are observed from layer 19 to 23 for both paths.
    }
    \vspace{-2.0mm}
    \label{fig:attn_layers}
\end{figure}

\begin{figure}[t]
    \centering
    \includegraphics[width=0.94\linewidth]{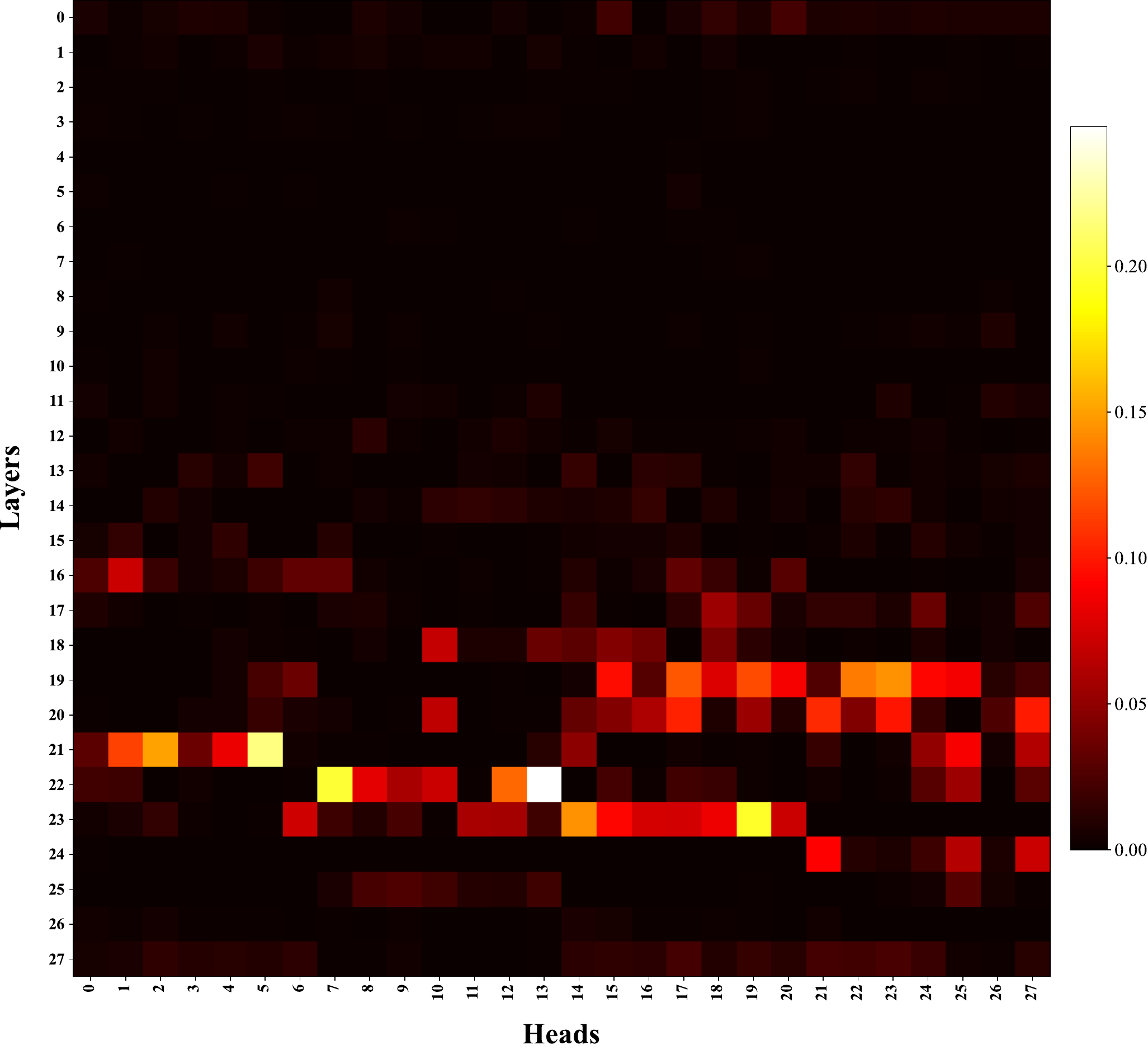}
    \vspace{-2pt}
    \caption{
    Heatmap of average attention to final items of partitions across layers and heads for intermediate counting steps. Layers and heads responsible for transferring numerical information from input partitions to intermediate counts appear with higher intensity.
    }
    \vspace{-2mm}

    \label{fig:attn_layers_heads}
\end{figure}

We analyze attention patterns of Qwen2.5-7B~\citep{qwen25} to understand how the proposed System-2 strategy enables large-scale counting. We first identify which tokens are attended to when the model generates (i) intermediate partition-level counts and (ii) the final aggregated answer. We then localize the layers and heads that contribute most to these processes.

Figure~\ref{fig:attn_reasoning} shows attention patterns to the input items when the model generates intermediate reasoning steps. We consider a structured context with three partitions containing 3, 4, and 6 items. During the generation of a reasoning token such as ``part 2: 4'', the attention weight of the generated number peaks sharply on the final item and the final comma separator of the corresponding partition. This pattern is consistent across prompts and is most pronounced in middle to late layers, typically around layers 19 to 23. Earlier items within the same partition receive substantially lower attention.

Figure~\ref{fig:attn_final} shows attention patterns when the model generates the final answer. The ``:'' token and the whitespace token immediately preceding the final number attend strongly to all intermediate reasoning numbers produced earlier (e.g., 3, 4, and 5). The attention mass is distributed across these tokens, with a clear concentration in the same layers identified for intermediate steps. Attention to the original input items is weak at this stage.

To further localize the attention pathways involved in intermediate reasoning and final aggregation, we focus on the most relevant tokens identified in Figure~\ref{fig:attention_system2}. We average attention values across different configurations, including item types and partition sizes. Figure~\ref{fig:attn_layers} shows that attention peaks in layers 19 to 23, highlighting the central role of these layers in information transfer and aggregation. In addition, Figure~\ref{fig:attn_layers_heads} further shows that the most influential heads are concentrated in layers 21, 22, and 23. For example, head 13 in layer 22 consistently exhibits high attention to the selected tokens.
Additional methodological details and attention analyses across multiple model architectures are provided in Appendix~\ref{sec:appendix_attention}.

Together, these results suggest a staged computation. First, each partition is counted independently, with the final tokens of the partition encoding the local count. Second, these local counts are written into intermediate reasoning tokens. Finally, the model attends to these intermediate tokens and aggregates their values to produce the final answer. These observations are consistent with a hypothetical mechanism that separates counting, information transfer, and aggregation into distinct components.

\section{Causal Mediation Analysis}
\label{sec:causal_analysis}

To assess the hypothesis of a multi-stage System-2 counting mechanism suggested by the attention analysis, we perform a set of causal mediation experiments based on activation patching~\citep{heimersheim2024activationpatching,zhang2024activationpatching_bestpractices}, masking ablation, and attention knockout~\citep{attention_knockout}. The goal is to identify where partition-level count information is stored, how it is transferred and consolidated, and which components are causally required for correct aggregation.

\subsection{Token-Level Information Probing}
\label{sec:token_probing}

\begin{figure}[t]
    \centering
    \includegraphics[width=0.94\linewidth]{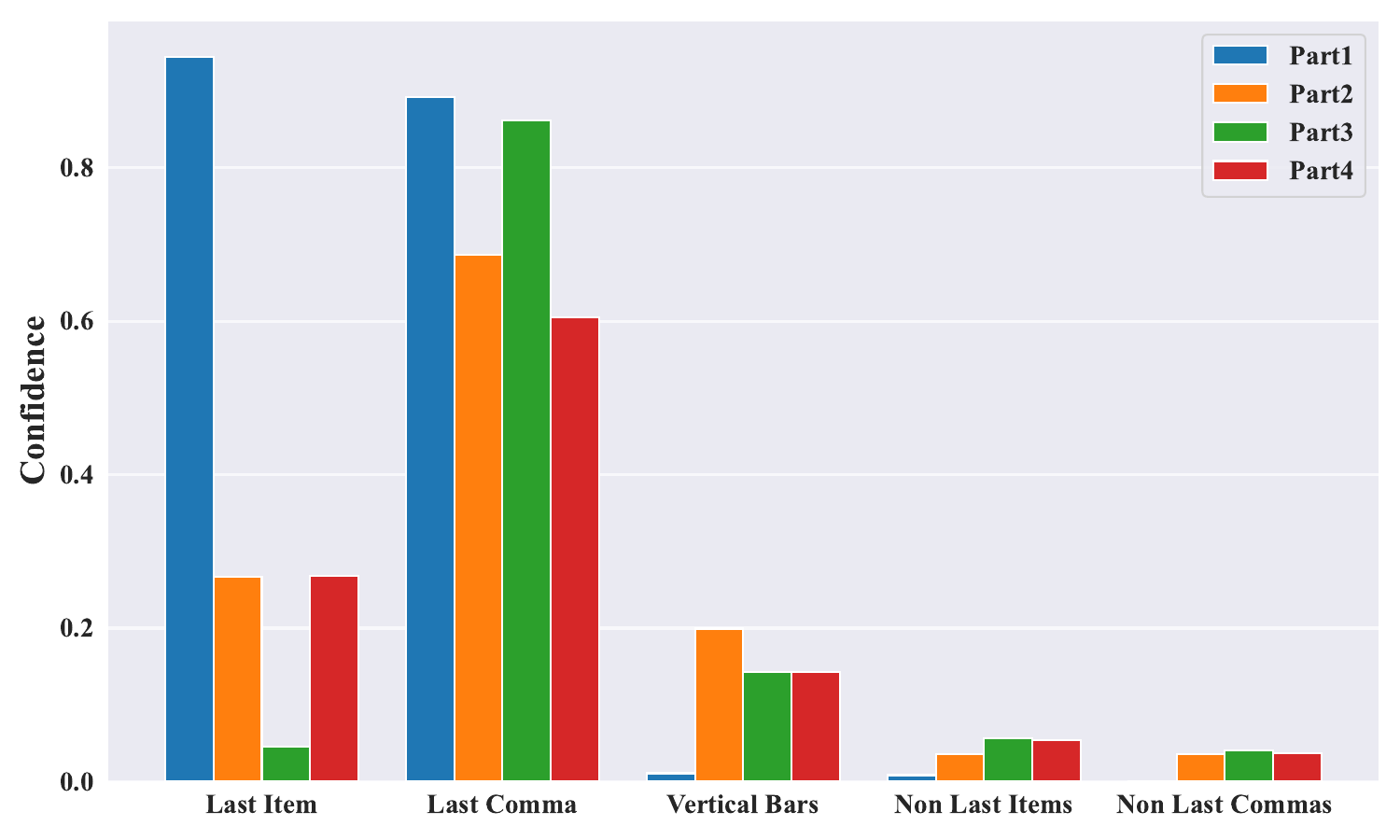}
    \vspace{-2pt}
    \caption{
    Ground-truth probabilities across different tokens of partitions decoded by CountScope.
    Probabilities are averaged over different item types and configurations (3 to 9 items per partition).
    }
    \label{fig:countscope_probs}
    \vspace{-2mm}
\end{figure}

We first examine where partition-level counts are encoded in the input context. This analysis requires vocabulary projection and probing tools. We observe that existing tools, such as logit-lens~\citep{nostalgebraist2020logitlens} and tuned-lens~\citep{belrose2023tunedlens}, are not reliable for decoding numerical information. We therefore use the CountScope method~\citep{hasani2025countscope} to decode the implicit count associated with tokens. CountScope is a causal probing method that patches the activation of a target token into a blank counting context and decodes the model’s output as the implied count.

Figure~\ref{fig:countscope_probs} shows the average decoding probability of ground-truth numbers across different token types. Ground-truth numbers correspond to the number of items in each partition. The results show that the ground-truth value is encoded with high confidence primarily at the final item and the final comma separator of each partition. For the first partition, the confidence at the final item is high; however, for subsequent partitions, the confidence at this token decreases, while the confidence at the vertical bar character increases. Nevertheless, the final comma separator reliably stores the latent count of the corresponding partition. These findings are consistent with the attention results in Section~\ref{sec:attention_analysis}, which showed that decoded numbers of intermediate steps attend most strongly to these final partition tokens.

More interestingly, this experiment reveals that at each partition boundary, the count resets and begins again. As a result, the final item of each partition encodes the number of items counted since the beginning of that partition.
This observation explains why the System-2 strategy avoids the large-number failure observed in unstructured inputs.
Each partition is counted independently, and its size remains within the range where the model’s implicit counter is accurate.

\subsection{Token-Level Causal Interventions}
\label{sec:token_ablation}

\begin{figure}[t]
    \centering
    \includegraphics[width=0.99\linewidth]{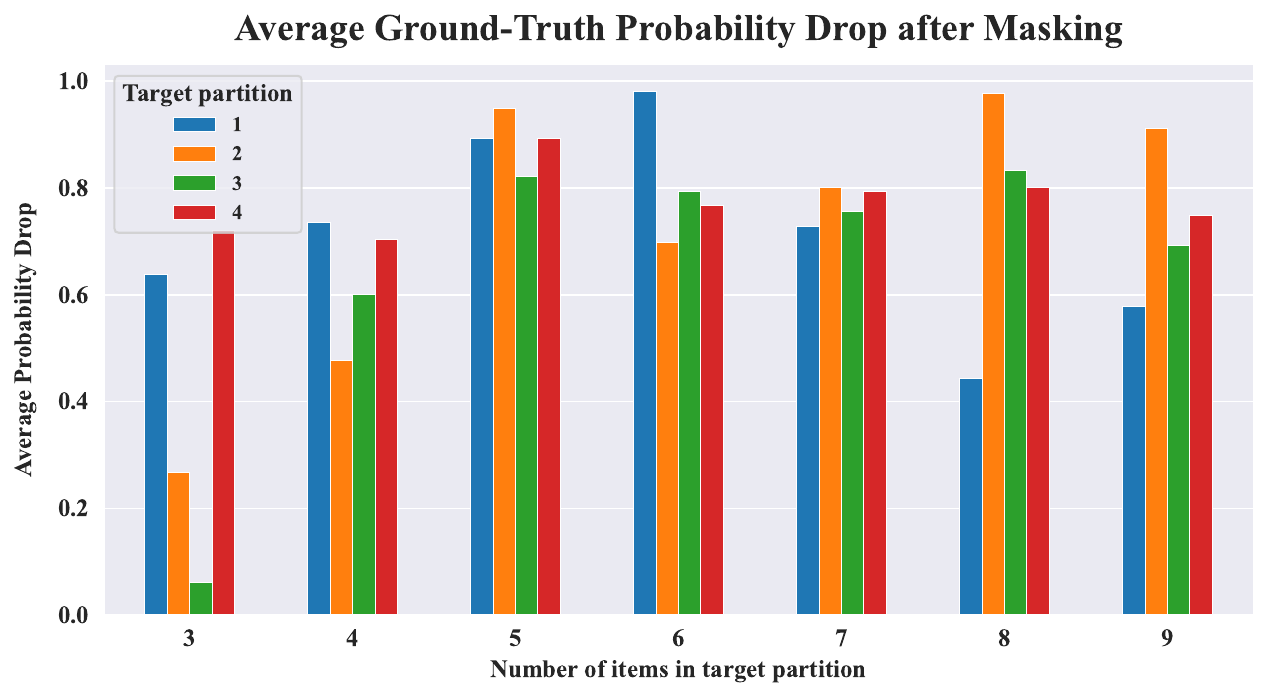}
    \vspace{-2pt}
    \caption{
    Average ground-truth probability drop after masking the final item and comma (e.g., `, apple') from each partition, showing the effect on the target count for digit sizes ranging from 3 to 9.
    }
    \vspace{-1.0mm}
    \label{fig:masking_prob_drop}
\end{figure}

\begin{figure*}[t]
    \centering
    \begin{subfigure}[t]{0.48\linewidth}
        \centering
        \includegraphics[width=\linewidth]{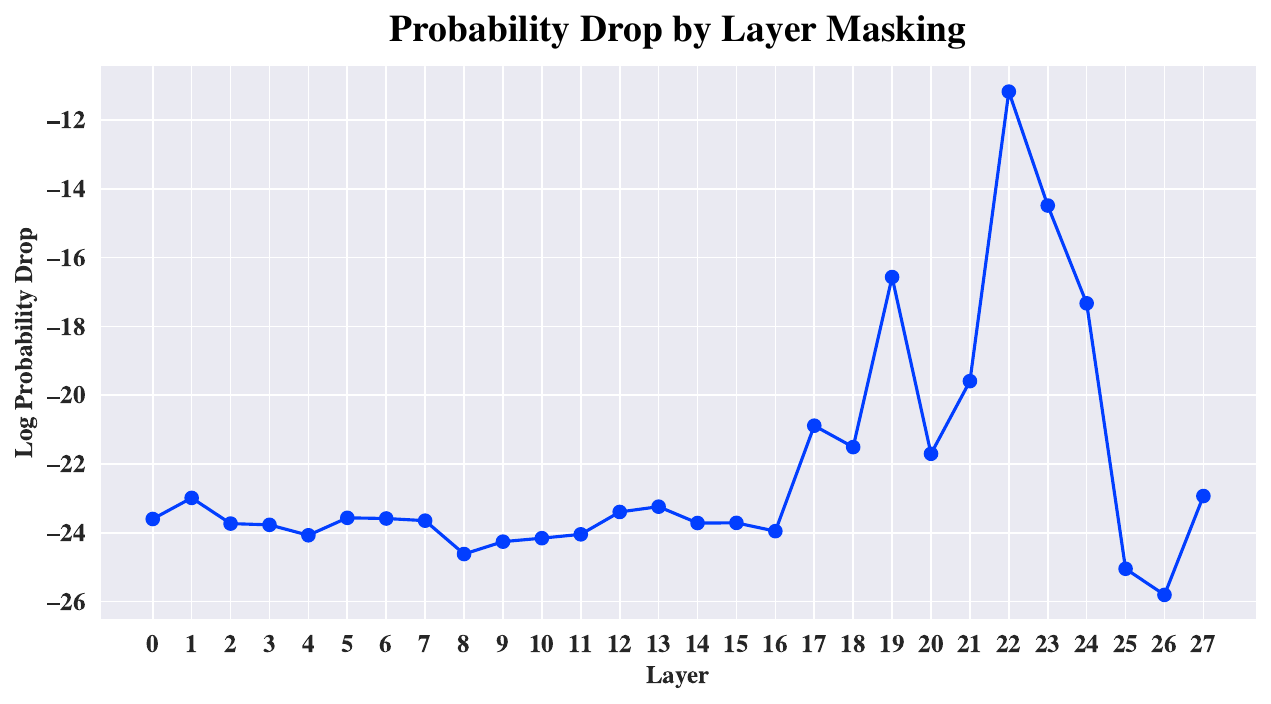}
        \caption{Layer masking (deletion)}
        \label{fig:layer_masking}
    \end{subfigure}
    \hfill
    \begin{subfigure}[t]{0.48\linewidth}
        \centering
        \includegraphics[width=\linewidth]{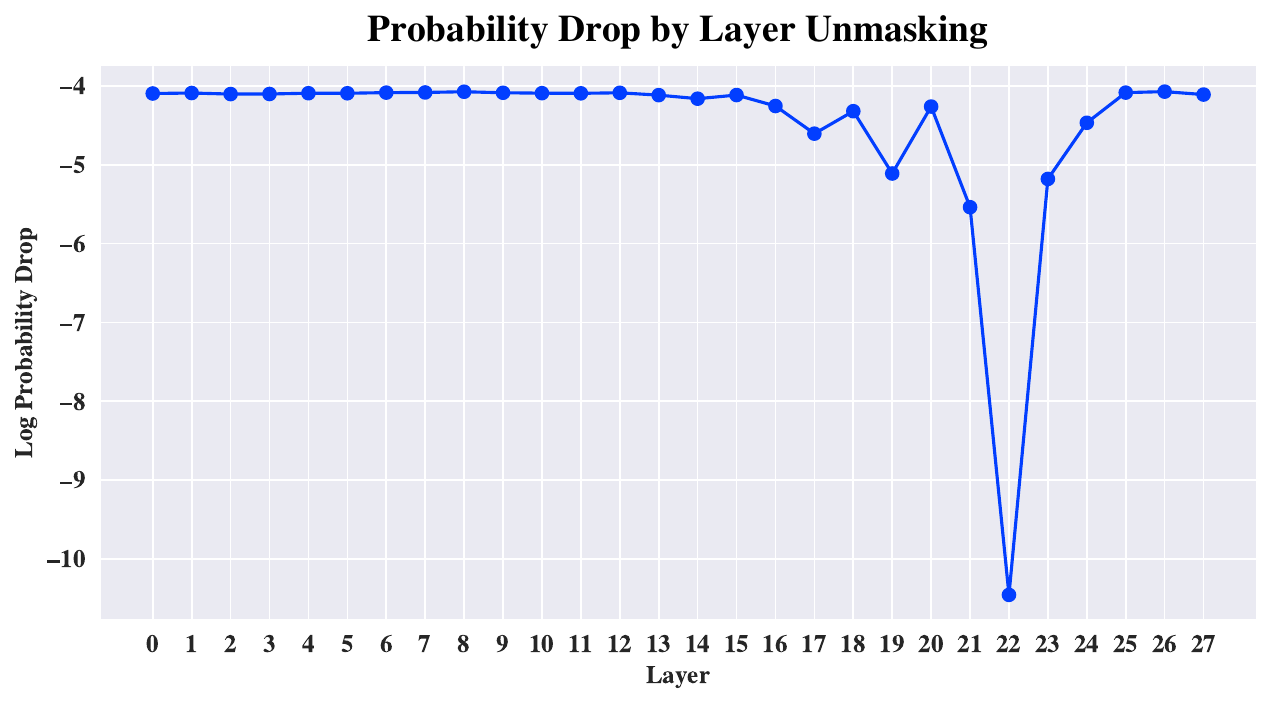}
        \caption{Layer unmasking (insertion)}
        \label{fig:layer_unmasking}
    \end{subfigure}
    \vspace{-2pt}
    \caption{Average change in log probability of intermediate counts after layer masking (deletion) and unmasking (insertion) of the final item and separator of each partition. In the masking experiment, embeddings from the target layer are zero-ablated. In the unmasking experiment, embeddings of the selected tokens are zero-ablated across all layers, and only the target-layer embeddings are restored from the clean run.}
    \label{fig:layer_localization}
    \vspace{-2mm}
\end{figure*}

To test whether the identified tokens causally mediate the model’s behavior, we perform zero ablation on the final items and separators of each partition. Specifically, after processing the entire input sequence and extracting token embeddings, we replace the activation values of selected tokens with zeros across all layers.

Figure~\ref{fig:masking_prob_drop} shows the average probability drop after zero ablation of the final item and final comma separator. This intervention leads to a sharp drop in the probability of generating the correct partition-level count in the corresponding reasoning step. For example, ablating the final ``, apple'' token of a four-item partition significantly reduces the likelihood of generating the token ``4''.

\subsection{Information Pathway Localization}
\label{sec:attention_knockout}

\begin{figure}[t]
    \centering
    \includegraphics[width=0.99\linewidth]{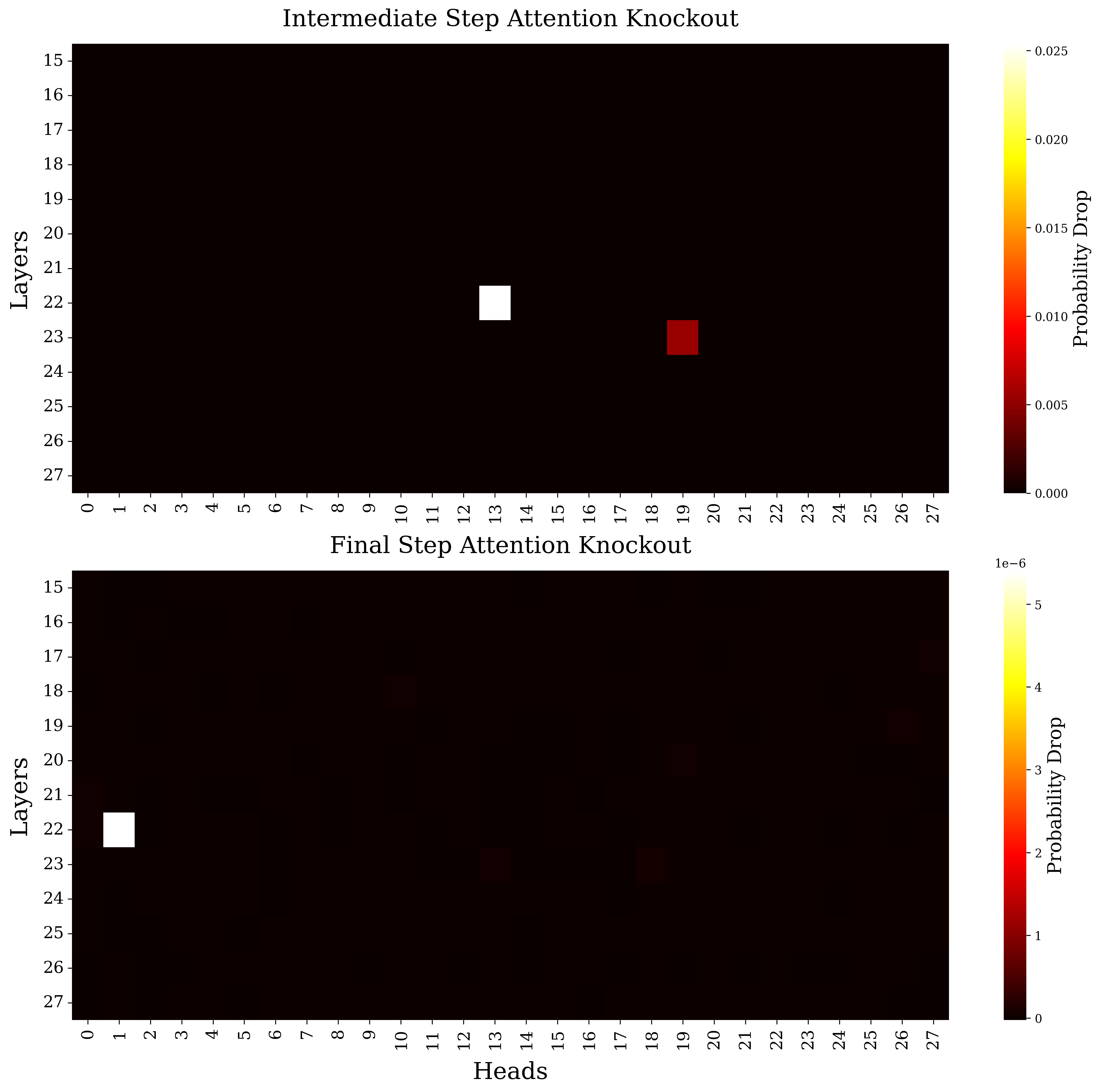}
    \vspace{-2pt}
    \caption{
    Heatmap of the average probability drop after attention knockout across layers and heads for intermediate (top) and final counting steps (bottom). Attention knockout is applied only to the identified effective tokens from the input context and intermediate steps.
    }
    \vspace{-1.0mm}
    \label{fig:attn_knockout}
\end{figure}

To perform fine-grained circuit localization, we analyze the pathways responsible for System-2 counting using attention knockout~\citep{attention_knockout}. We selectively block individual attention heads and layers and measure the resulting drop in counting accuracy for intermediate and final steps. 
Figure~\ref{fig:layer_localization} shows that for Qwen2.5-7B, the most influential components are concentrated in layer 22. In particular, attention head 13 is most important for intermediate steps, while head 1 is most important for the final aggregation step (Figure~\ref{fig:attn_knockout}).

Notably, head 13 in layer 22 also exhibits the highest attention values in the attention heatmap shown in Figure~\ref{fig:attn_layers_heads}. Compared to Figure~\ref{fig:attn_layers_heads}, the pattern in Figure~\ref{fig:attn_knockout} is considerably sparser. A plausible explanation is the presence of parallel information pathways, where knocking out a single pathway does not fully disrupt the computation because other pathways can partially compensate.
Finally, we observe that the heads responsible for transferring partition-level information from the input context to the intermediate steps are not necessarily the same as those responsible for transferring information from the intermediate reasoning steps to the final answer, even when they are located in the same layer.

\subsection{Cross-Context Activation Patching}
\label{sec:cross_context}

Finally, we perform cross-context activation patching to test how partition-level information is combined to form the final answer. We focus on intermediate responses and final aggregation steps. Our attention and masking analyses indicate that transferred numerical information from the context to intermediate steps is consolidated in the specific tokens ``:'', the whitespace token, and the partition number. Here, we investigate this behavior using a causal intervention setup~\citep{geiger2021causalabstractions}.

To this end, we sample two different contexts, each containing four partitions, with partition sizes randomly chosen between 3 and 9. After generating the intermediate steps, we select one partition and transfer the embeddings from layers 18 to 24 of the target tokens between the two responses. 
For example, let ``part 1: 7, part 2\textcolor{blue}{: 4}, part 3: 8'' be the intermediate steps of the first context and ``part 1: 5, part 2\textcolor{blue}{: 6}, part 3: 3'' be those of the second context. We swap the intermediate activations of the tokens highlighted in blue between the two responses.
After this intervention, we allow the model to generate the final response.

\begin{figure}[t]
    \centering
    \includegraphics[width=0.99\linewidth]{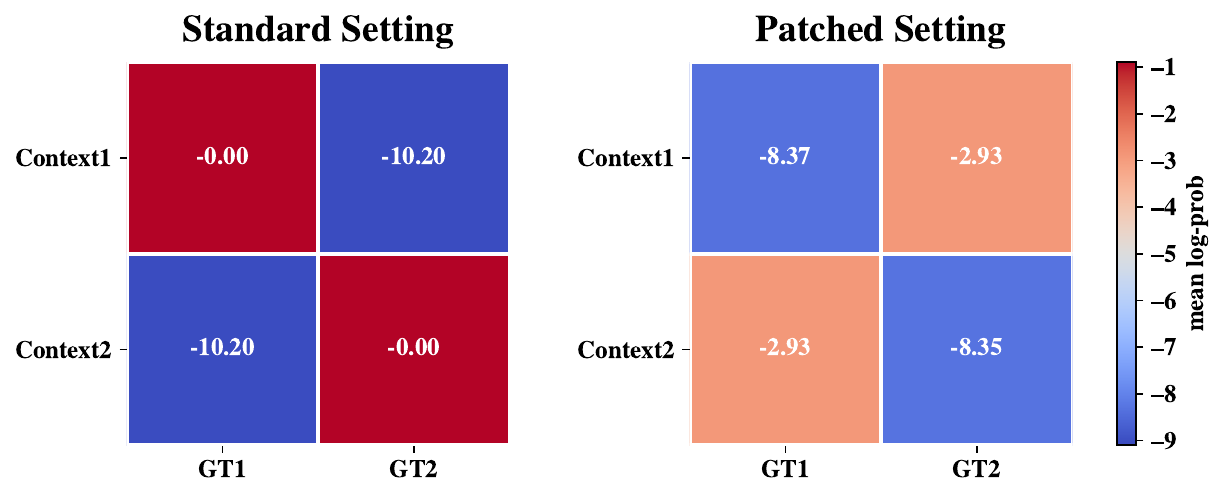}
    \vspace{-4pt}
    \caption{
    Average log probabilities of the first and second contexts before (left) and after (right) activation patching. The first and second columns correspond to the predicted outputs (total sums) of the first and second contexts. After swapping the layer embeddings of the selected tokens from a given partition between the two contexts, the output no longer follows the original total sum and instead reflects the transferred number. Values are averaged across different configurations (selected partitions, partition sizes, and item types).
    }
    \vspace{-2.0mm}
    \label{fig:activation_patching_logprobs}
\end{figure}

This manipulation causally affects the corresponding intermediate steps and changes the final count accordingly. In the given example, the total sums of the first and second contexts are 19 and 14 before activation patching. After swapping the numerical contents of their second steps (shown in blue), the total for the first context becomes 21 and the total for the second context becomes 12. This shows that the final sum is causally mediated by the intermediate-layer embeddings of the tokens ``:'', whitespace, and the partition number.
Figure~\ref{fig:activation_patching_logprobs} shows the log probabilities for this experiment, averaged over various configurations.

\section{Related Work}
\label{sec:related}

\paragraph{Counting in LLMs.}
A growing body of work shows that counting in LLMs is brittle: performance can hinge on surface form and segmentation rather than a stable procedure \citep{fu2024lettercounting}. Subword tokenization can blur item boundaries and measurably affect counting accuracy \citep{zhang2024countingabilitytokenization}. More broadly, deterministic-task evaluations (including counting) can flip under minor prompt/content changes, complicating extrapolation from narrow setups \citep{ball2024fixed_effect_fallacy}. On the theory side, transformers can exactly count token frequencies only in specific regimes, with feasibility tied to representation capacity and context-length scaling \citep{yehudai2024whencantransformerscount}.

\paragraph{Numerical representations and mechanistic views.}
Cognitive work argues for specialized number systems and structured magnitude representations \citep{feigenson2004core,dehaene2011numbersense}, motivating analogous questions in neural models. Representation analyses study how numerical magnitude is organized in LLM hidden spaces and relate these structures to human-like effects \citep{alquboj2025numberrepresentations}. Mechanistic studies localize how counting signals emerge and update across items in controlled settings \citep{golkar2024contextualcounting}. Closest to our setting, \citet{hasani2025countscope} combine a target-context probe with layerwise/token-level analyses and causal interventions to identify latent counter-like signals across both LLMs and LVLMs.

\paragraph{Decomposition and reasoning traces.}
Chain-of-Thought can improve multi-step reasoning by eliciting intermediate computations \citep{wei2022cot}, but explicit traces are not necessarily faithful. Question decomposition work studies when decomposition improves faithfulness rather than mainly serving as scaffolding \citep{radhakrishnan2023questiondecomposition}. Decomposing images into smaller regions has also been shown to be effective for visual reasoning \citep{izadi2025viser, grounding_id}. This strategy improves visual counting performance \citep{izadi2025viser,qharabagh2024lvlmcount, grounding_id}, but its underlying mechanism remains underexplored.

\paragraph{Interpretability tools.}
Causal interventions such as activation patching help identify computation-critical states \citep{heimersheim2024activationpatching,geiger2021causalabstractions}, with best-practice guidance for reliable metrics and methods \citep{zhang2024activationpatching_bestpractices}. Patch-based inspection frameworks further systematize intervention/inspection configurations \citep{ghandeharioun2024patchscopes}. Readout-style probes provide a complementary perspective by tracking when information becomes linearly decodable across layers, including the Logit Lens \citep{nostalgebraist2020logitlens} and learned variants such as the Tuned Lens \citep{belrose2023tunedlens}.

\paragraph{Positioning of This Work.}
Our setup is closest to System-2-style prompting via intermediate steps: like CoT, we elicit explicit intermediate computations \citep{wei2022cot}, and like task decomposition, we solve the instance by breaking it into sequential subproblems \citep{radhakrishnan2023questiondecomposition}. The key difference is that the decomposition is tied to a \emph{structured input format}: following prior vision-based studies \citep{izadi2025viser, grounding_id}, we first partition the input text into marked segments and then use a CoT protocol aligned with this structure by producing segment-level subcounts and then aggregating them. This creates explicit stage boundaries that are absent in monolithic counting prompts and lets us study how counting representations evolve across stages. Mechanistically, our analysis aligns with prior work that localizes latent counter-like signals and tests them with interventions \citep{hasani2025countscope}, while shifting the question to how the \emph{structured CoT decomposition} reshapes internal computation on long contexts. In this way, we connect behavioral brittleness observations \citep{ball2024fixed_effect_fallacy,zhang2024countingabilitytokenization} to a stage-wise mechanistic account of how count information is formed, routed, and combined under structured prompting \citep{heimersheim2024activationpatching,zhang2024activationpatching_bestpractices}.

\section{Conclusion}
\label{sec:conclusion}

This paper showed that the failure of large language models on large-scale counting tasks arises from reliance on System-1–like implicit counting mechanisms with limited capacity~\citep{kahneman2011thinking,zhang2024countingabilitytokenization}. This problem reflects architectural constraints of transformer models rather than fundamental limits of numerical reasoning. We introduced a simple System-2 test-time strategy that decomposes large counting problems into smaller sub-tasks and aggregates their results, enabling accurate counting over long contexts without modifying model parameters or training procedures.

In addition to behavioral improvements, we provided a mechanistic explanation of how this strategy operates internally. Using attention analysis and causal mediation methods~\citep{hasani2025countscope,heimersheim2024activationpatching}, we showed that partition-level counts are encoded at boundary tokens, transferred through specific attention pathways to intermediate reasoning steps, and aggregated in middle-to-late layers to form the final answer. Intermediate reasoning tokens play a crucial role in this process, mediating the flow of numerical information from input partitions to the final output.

In principle, this procedure removes a fixed upper bound on countable size, as long as each sub-problem remains within the model’s reliable regime.
While our experiments focus on counting, the same analysis framework may extend to other reasoning tasks where implicit representations saturate and explicit decomposition into sub-tasks facilitates correct behavior.
This study highlights how structured test-time strategies can reveal and extend the computational abilities of existing models, offering a path toward both improved performance and deeper interpretability.
Future research could explore further applications of this strategy to more complex reasoning tasks in language and beyond multimodal settings.


\section*{Limitations}
\label{sec:limitations}

This work focuses on a text-based counting task with limited object diversity. For systematic mechanistic analysis, we use synthetic contexts with repeated nouns rather than natural free-form text. The approach depends on structured prompts and assumes prior knowledge of the model’s reliable counting range. While this requirement is mild in practice, it may limit applicability across domains. The proposed decomposition strategy is most effective for tasks that can be divided into near-independent sub-tasks (e.g., multi-step arithmetic or sequential planning). It is not expected to generalize to tasks with strong interdependencies, where subproblems cannot be cleanly separated.

\bibliography{custom}

\clearpage
\pagebreak

\appendix

\section{Appendix}
\label{sec:appendix}

\newtcolorbox{promptbox}[2][]{
  colback=gray!3,
  colframe=gray!50,
  coltitle=white,
  fonttitle=\bfseries,
  title={#2},
  boxsep=4pt,
  left=4pt,
  right=4pt,
  top=4pt,
  bottom=4pt,
  colbacktitle=black,
  #1
}

\subsection{Details of the Experimental Setup}
\label{sec:behavioral_setup}

We evaluate counting performance under controlled variations of input format and prompting strategy. Each example consists of a list of repeated single-word items, and the model is required to output the total count. Following the evaluation protocol summarized in Tables~\ref{tab:behavioral:open_source} and~\ref{tab:behavioral:closed_source}, we vary (i) whether the input is \emph{structured} or \emph{unstructured}, and (ii) whether the model is explicitly encouraged to produce intermediate reasoning steps (\emph{with steps} vs.\ \emph{without steps}). Performance is reported using exact-match accuracy and mean absolute error (MAE) across different context-length ranges.

\paragraph{Item types.}
All inputs are constructed from simple, common nouns drawn from two semantic categories: \emph{fruits} and \emph{animals}. The complete candidate set is shown below.
\begin{promptbox}{Items}
\texttt{apple, orange, peach, fig, mango, pear, coconut, cherry, plum,}\\
\texttt{cat, dog, horse, rabbit, whale, cow, frog}
\end{promptbox}

\paragraph{Input formats.}
In the \textbf{unstructured} setting, the input is a flat, comma-separated list of $n$ identical items. In the \textbf{structured} setting, the same multiset is divided into multiple partitions separated by a vertical bar (\texttt{|}), enabling explicit local counting followed by summation.

\paragraph{Partition sizes.}
For \textbf{open-source models} (Table~\ref{tab:behavioral:open_source}), each partition contains between \textbf{6 and 9 items}. For \textbf{closed-source models} (Table~\ref{tab:behavioral:closed_source}), which are evaluated on larger context lengths, partition sizes range from \textbf{15 to 25 items}. In all cases, partitions are constructed such that their sizes sum exactly to the target length $n$.

\paragraph{Prompting strategies.}
For each input format, we evaluate two prompting strategies: \emph{without intermediate steps} and \emph{with intermediate steps}. These settings correspond directly to the ``w/o steps'' and ``w/ steps'' rows reported in Tables~\ref{tab:behavioral:open_source} and~\ref{tab:behavioral:closed_source}.

\begin{promptbox}{Unstructured Input (w/o steps)}
\texttt{You will be given a list of items. Count the total number of objects.}\\
\texttt{Output the final result exactly in the following format:}\\
\texttt{Final answer: [x]}
\end{promptbox}

\begin{promptbox}{Unstructured Input (w/ steps)}
\texttt{You will be given a list of items. Count the total number of objects.}\\
\texttt{Let's count step by step.}\\
\texttt{Output the final result exactly in the following format:}\\
\texttt{Final answer: [x]}
\end{promptbox}

\begin{promptbox}{Structured Input (w/o steps)}
\texttt{You will be given multiple partitions of content separated by ``|''.}\\
\texttt{For each partition, count the number of items it contains.}\\
\texttt{After counting all partitions, compute the total by summing the counts.}\\
\texttt{Output the final result exactly in the following format:}\\
\texttt{Final answer: [x]}
\end{promptbox}

\begin{promptbox}{Structured Input (w/ steps)}
\texttt{You will be given multiple partitions of content separated by ``|''.}\\
\texttt{For each partition:}\\
\texttt{- Count the number of items it contains.}\\
\texttt{- Report the count separately using the format:}\\
\texttt{  part1: [x1]}\\
\texttt{  part2: [x2]}\\
\texttt{  ...}\\
\texttt{After counting all partitions:}\\
\texttt{- Compute the total by summing all individual counts.}\\
\texttt{- Output the final total exactly in this format:}\\
\texttt{Final answer: [x]}
\end{promptbox}

\begin{table*}[htb]
\centering
\resizebox{0.68\linewidth}{!}{
\begin{tabular}{lcc|cccc}
\hline
\textbf{Model} & \textbf{Input} & \textbf{Output} 
& \textbf{11--20} & \textbf{21--30} & \textbf{31--40} & \textbf{41--50} \\
\hline

\multirow{4}{*}{\textbf{Gemma2-2B}}
& \multirow{2}{*}{Unstructured} & w/o steps
& 0.01 & 0.00 & 0.00 & 0.00 \\
&  & w/ steps
& 0.00 & 0.00 & 0.00 & 0.00 \\
\cline{2-7}
& \multirow{2}{*}{Structured} & w/o steps
& 0.05 & 0.12 & 0.03 & 0.04 \\
&  & w/ steps
& \textbf{0.18} & \textbf{0.15} & \textbf{0.10} & \textbf{0.07} \\
\hline

\multirow{4}{*}{\textbf{Gemma3-4B}}
& \multirow{2}{*}{Unstructured} & w/o steps
& 0.91 & 0.45 & 0.04 & 0.08 \\
&  & w/ steps
& \textbf{0.99} & 0.52 & 0.02 & 0.03 \\
\cline{2-7}
& \multirow{2}{*}{Structured} & w/o steps
& 0.26 & 0.13 & 0.03 & 0.05 \\
&  & w/ steps
& 0.96 & \textbf{0.56} & \textbf{0.19} & \textbf{0.17} \\
\hline

\end{tabular}
}
\caption{Average accuracy on \textbf{smaller open-source models} (Gemma2-2B and Gemma3-4B) for context sizes from 11 to 50. Each model is evaluated on structured and unstructured inputs, with and without intermediate reasoning steps.}
\label{tab:appendix_small_models}
\vspace{-1mm}
\end{table*}
\begin{table*}[htb]
\centering
\resizebox{0.73\linewidth}{!}{
\begin{tabular}{lcc|cccc}
\hline
\textbf{Model} & \textbf{Input} & \textbf{Output} 
& \textbf{11--20} & \textbf{21--30} & \textbf{31--40} & \textbf{41--50} \\
\hline

\multirow{4}{*}{\textbf{Qwen2.5-Math-7B}}
& \multirow{2}{*}{Unstructured} & w/o steps
& 0.95 & 0.54 & 0.04 & 0.04 \\
&  & w/ steps
& 0.94 & 0.46 & 0.02 & 0.05 \\
\cline{2-7}
& \multirow{2}{*}{Structured} & w/o steps
& 0.97 & \textbf{0.89} & \textbf{0.89} & \textbf{0.75} \\
&  & w/ steps
& \textbf{1.00} & 0.77 & 0.75 & 0.52 \\
\hline

\multirow{4}{*}{\textbf{DeepSeekMath-7B}}
& \multirow{2}{*}{Unstructured} & w/o steps
& 0.86 & 0.12 & 0.04 & 0.04 \\
&  & w/ steps
& 0.90 & 0.30 & 0.03 & 0.05 \\
\cline{2-7}
& \multirow{2}{*}{Structured} & w/o steps
& 0.40 & 0.21 & 0.03 & 0.01 \\
&  & w/ steps
& \textbf{0.86} & \textbf{0.58} & \textbf{0.38} & \textbf{0.19} \\
\hline

\end{tabular}
}
\caption{Average accuracy on \textbf{math-specialized models} for context sizes from 11 to 50. DeepSeekMath-7B follows the same trend as general-purpose models, where structured input with intermediate reasoning performs best. Qwen2.5-Math-7B shows stronger performance for both structured settings, suggesting improved internal aggregation of partition-level counts.}
\label{tab:appendix_math_models}
\vspace{-1mm}
\end{table*}

\subsection{Behavioral Results on Additional Models}
\label{sec:appendix_additional_models}

We extend the behavioral evaluation to two smaller open-source models, Gemma2-2B~\citep{gemma2} and Gemma3-4B~\citep{gemma3}, and two math-specialized models, Qwen2.5-Math-7B~\citep{qwen2math} and DeepSeekMath-7B~\citep{deepseekmath}. We use the same four settings as in the main paper: unstructured vs. structured inputs, each with short-answer generation or intermediate reasoning steps. We evaluate context sizes from 11 to 50 items, grouped into four bins.

Results for smaller models are shown in Table~\ref{tab:appendix_small_models}. As expected, absolute accuracy decreases with model size, especially for longer contexts. However, the main trend from the paper remains unchanged. Structured input with intermediate reasoning is the strongest setting overall and remains more robust than the other three variants as context length increases. This suggests that the proposed decomposition strategy is not limited to larger models and still provides benefits in lower-capacity regimes.

Results for math-specialized models are shown in Table~\ref{tab:appendix_math_models}. DeepSeekMath-7B follows the same pattern as general-purpose models, where structured input with intermediate reasoning gives the best performance. Qwen2.5-Math-7B shows a different behavior. Both structured settings perform strongly, and structured input without intermediate reasoning is often competitive with, or better than, the version with reasoning steps. This suggests that some math-tuned models may better combine partition-level counts internally without requiring explicit intermediate outputs. This behavior is also consistent with stronger out-of-context reasoning ability, where some models remain effective in the no-CoT regime~\citep{deng2311implicit, yao2025language}.

These additional experiments support the main conclusion that external structure is highly useful for long-range counting. For most models, the strongest configuration remains structured input with intermediate reasoning. At the same time, the Qwen2.5-Math result suggests that domain-specific fine-tuning can partially change how aggregation is carried out at inference time while maintaining strong performance in the no-CoT setting.

\subsection{Robustness to Tokenization and Input Structure}
\label{sec:robustness_input_structure}

We evaluate the generality of our System-2 counting mechanism across different tokenization schemes and input formats. Table~\ref{tab:tokenizers} summarizes the tokenizer types and vocabulary sizes of the models used in our main experiments. Despite these differences, the overall behavioral and mechanistic patterns are consistent, indicating that the counting strategy does not rely on a specific tokenizer or vocabulary.

\begin{table}[htb]
\centering
\resizebox{0.999\linewidth}{!}{
\begin{tabular}{lcc}
\hline
\textbf{Model} & \textbf{Tokenizer Type} & \textbf{Vocabulary Size} \\
\hline
GPT-4o & BPE & $\sim$200k \\
Gemini 2.5 Pro & SentencePiece & $\sim$256k \\
Gemma 3 & SentencePiece & $\sim$262k \\
Llama 3 & BPE & $\sim$128k \\
Qwen 2.5 & Byte-level BPE & $\sim$150k \\
\hline
\end{tabular}
}
\caption{Tokenizer types and vocabulary sizes for the evaluated models.}
\label{tab:tokenizers}
\end{table}

To further assess robustness, we evaluate two alternative input structures with different separators and instructions:

\textbf{Structure A}: Intermediate structures use part labels with commas separating items, e.g., 
\begin{lstlisting}
part1: apple, apple
part2: apple, apple, apple
\end{lstlisting}

\textbf{Structure B}: Intermediate structures list items with spaces, and parts are separated by slashes, e.g.,  
\begin{lstlisting}
apple apple / apple apple apple
\end{lstlisting}

Table~\ref{tab:appendix_structures} shows that the same performance trends hold for Qwen2.5-7B: structured inputs combined with intermediate reasoning steps consistently outperform unstructured inputs or inputs without reasoning steps, confirming that the System-2 mechanism is robust to minor changes in input formatting. However, the superior performance of Structure A over Structure B highlights the role of input structure in overall performance.

\begin{table*}[htb]
\centering
\resizebox{0.68\linewidth}{!}{
\begin{tabular}{lcc|cccc}
\hline
\textbf{Structure} & \textbf{Input} & \textbf{Output} & \textbf{11--20} & \textbf{21--30} & \textbf{31--40} & \textbf{41--50} \\
\hline
\multirow{4}{*}{\textbf{A}} 
& Unstructured & w/o steps & 0.30 & 0.11 & 0.05 & 0.00 \\
& Unstructured & w/ steps & 0.45 & 0.10 & 0.02 & 0.00 \\
\cline{2-7}
& Structured & w/o steps & 0.38 & 0.24 & 0.16 & 0.10 \\
& Structured & w/ steps & \textbf{1.00} & \textbf{0.77} & \textbf{0.64} & \textbf{0.32} \\
\hline
\multirow{4}{*}{\textbf{B}} 
& Unstructured & w/o steps & 0.28 & 0.09 & 0.05 & 0.01 \\
& Unstructured & w/ steps & 0.28 & 0.10 & 0.04 & 0.01 \\
\cline{2-7}
& Structured & w/o steps & 0.22 & 0.19 & 0.12 & 0.10 \\
& Structured & w/ steps & \textbf{0.74} & \textbf{0.46} & \textbf{0.39} & \textbf{0.24} \\
\hline
\end{tabular}
}
\caption{Accuracy for Qwen2.5 7B under two different input formats (A and B) with structured and unstructured inputs, with and without intermediate reasoning steps.}
\label{tab:appendix_structures}
\end{table*}

\subsection{Attention Analysis}
\label{sec:appendix_attention}
\paragraph{Input Details}

The exact prompt structure is provided below:

\begin{lstlisting}
You will be given a list of items, where groups (partitions) are separated by the "|" character.

For each partition:
- Count the number of items it contains.
- Report the count separately using the format:
  part1: x1
  part2: x2
  part3: x3
  ...

After counting all partitions:
- Compute the total by summing all individual counts.
- Output the final total exactly in this format:
Final answer: x
Just answer in this format without any extra things and follow the instructions.

apple, apple, apple | apple, apple, apple, apple | apple, apple, apple, apple, apple
\end{lstlisting}

To evaluate robustness across varying input structures, the final list provided to the model is randomized across experiments.  Additionally, to reduce potential token-specific biases, we vary the item labels used in the lists, drawing from the Appendix~\ref{sec:behavioral_setup}

\paragraph{Further Attention Results}
We extend the attention analysis to several large language models, including Qwen2.5-7B~\citep{qwen25}, Llama3.2 8B~\citep{llama3}, and Gemma3-4B~\citep{gemma3}. These models differ in architectural depth and layer organization, enabling a more comprehensive comparative analysis of attention behaviors across architectures.

\begin{figure*}[t]
    \centering
    \begin{subfigure}[t]{0.45\linewidth}
        \centering
        \includegraphics[width=\linewidth]{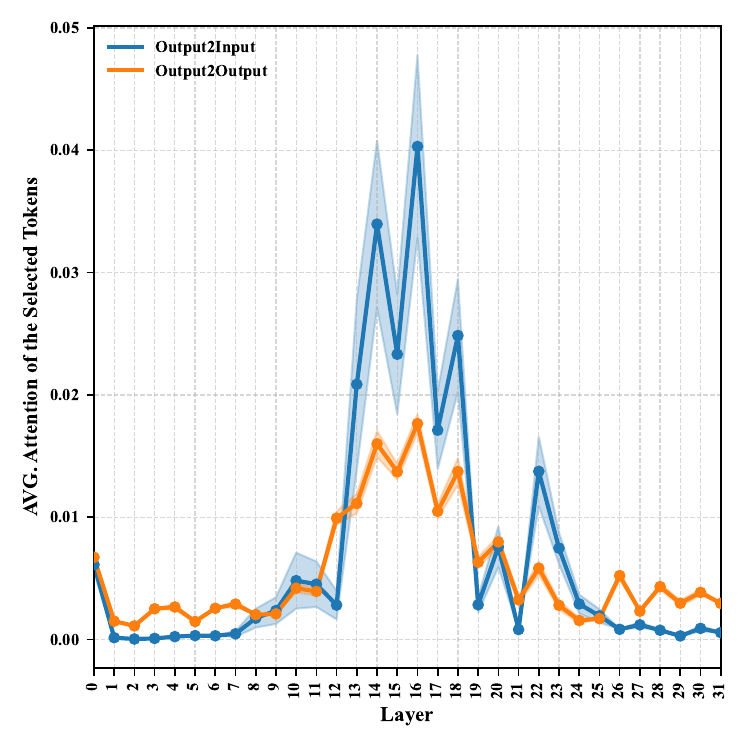}
        \caption{Average attention weights across layers for Llama3.2 8B. We report attention from selected output tokens to both input and output tokens. Attention to the key partition-related tokens peaks consistently between layers 13 and 18, motivating our choice of this interval for full attention visualization.}
        \label{fig:llama_layers}
    \end{subfigure}
    \hfill
    \begin{subfigure}[t]{0.45\linewidth}
        \centering
        \includegraphics[width=\linewidth]{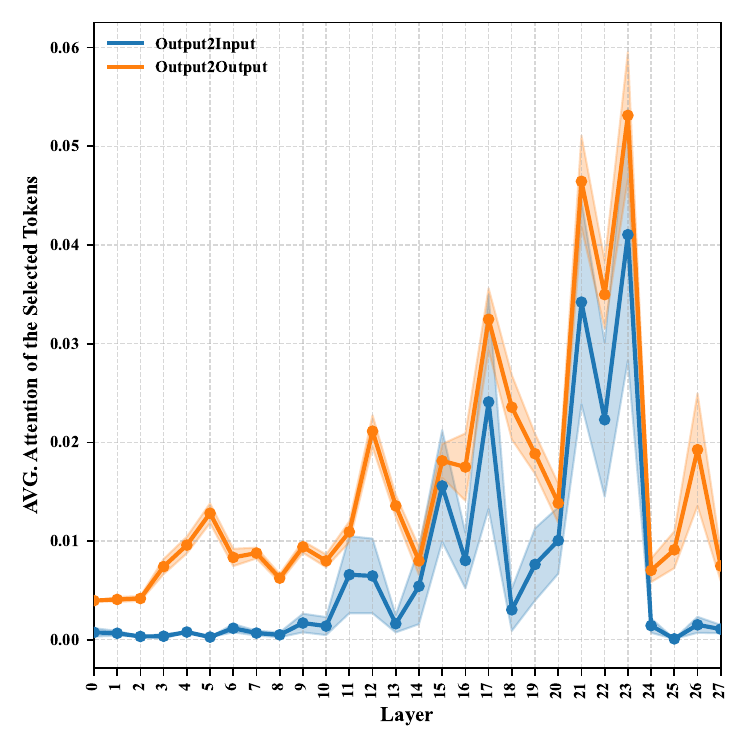}
        \caption{Average attention weights across layers for Gemma3-4B. We report attention from selected output tokens to both input and output tokens (as defined in Figure~\ref{fig:attn_reasoning}). Attention magnitude peaks between layers 21 and 23.}
        \label{fig:gemma_layers}
    \end{subfigure}
    \caption{Layer-wise attention analysis for Llama3.2-8B and Gemma3-4B. For each model, we visualize the average attention from selected output tokens to salient input tokens (notably the final item and comma of the partition). These trends are used to identify the layer ranges with the strongest partition-relevant attention.}
    \label{fig:other_model_layers}
\end{figure*}

As shown in Figure~\ref{fig:other_model_layers}, we analyze attention scores across different layer ranges for each model. For Gemma3-4B, we examine layers 21 to 23, while for Llama3.2-8B we consider layers 13 to 18. These layer intervals are selected based on the average attention magnitude from output tokens to key input tokens (specifically, the last item and the trailing comma of the predicted partition), which consistently peak within these ranges.

\begin{figure*}[t]
    \centering
    \includegraphics[width=0.9\linewidth]{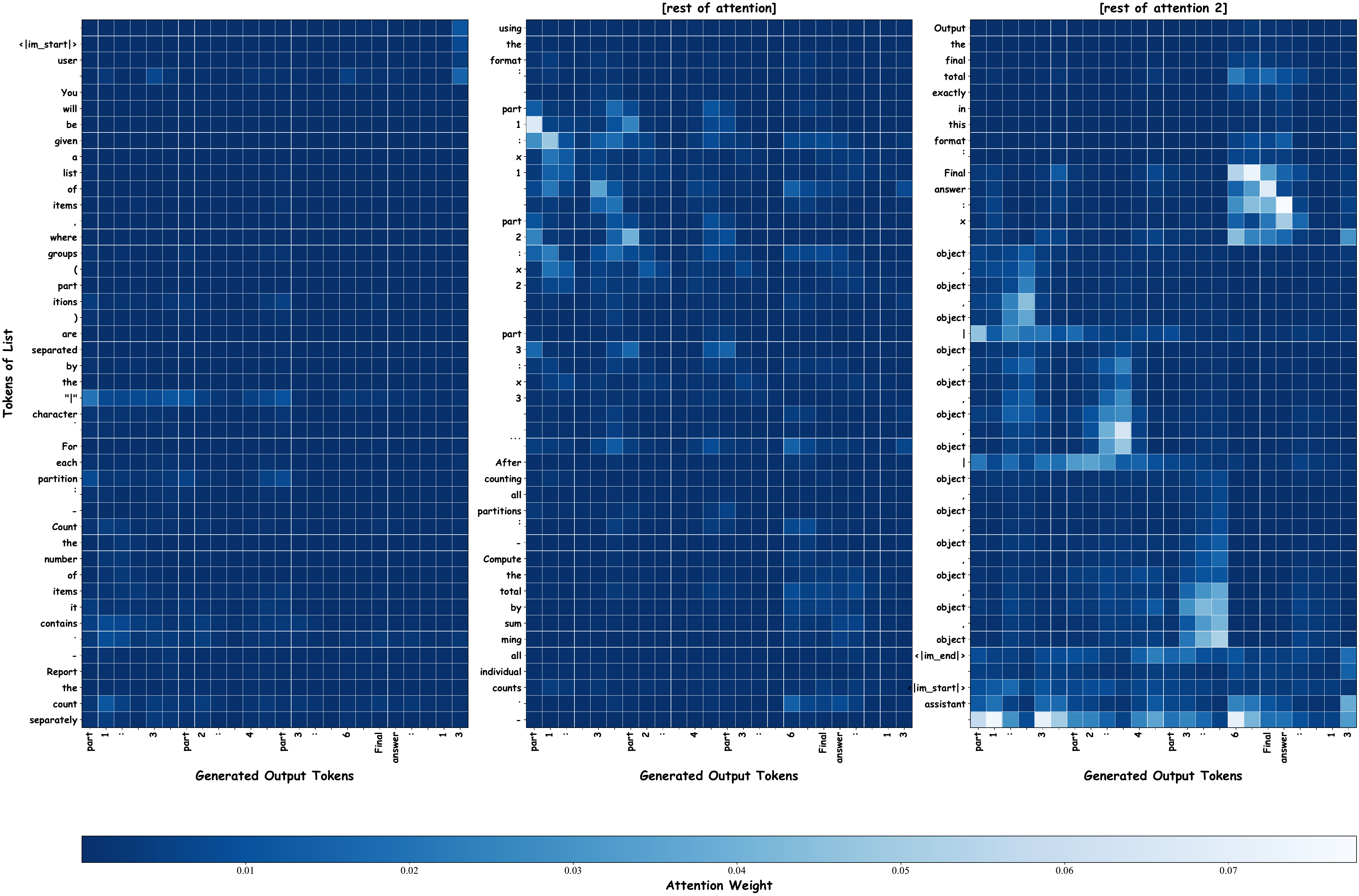}
    \caption{Full attention map for Qwen2.5-7B. The x-axis corresponds to output tokens and the y-axis to input tokens. The token immediately preceding the generated number of parts exhibits the strongest attention toward the final item and trailing comma of each partition matches its own segment.}
    \label{fig:qwen_full}
\end{figure*}

\begin{figure*}[t]
    \centering
    \includegraphics[width=0.9\linewidth]{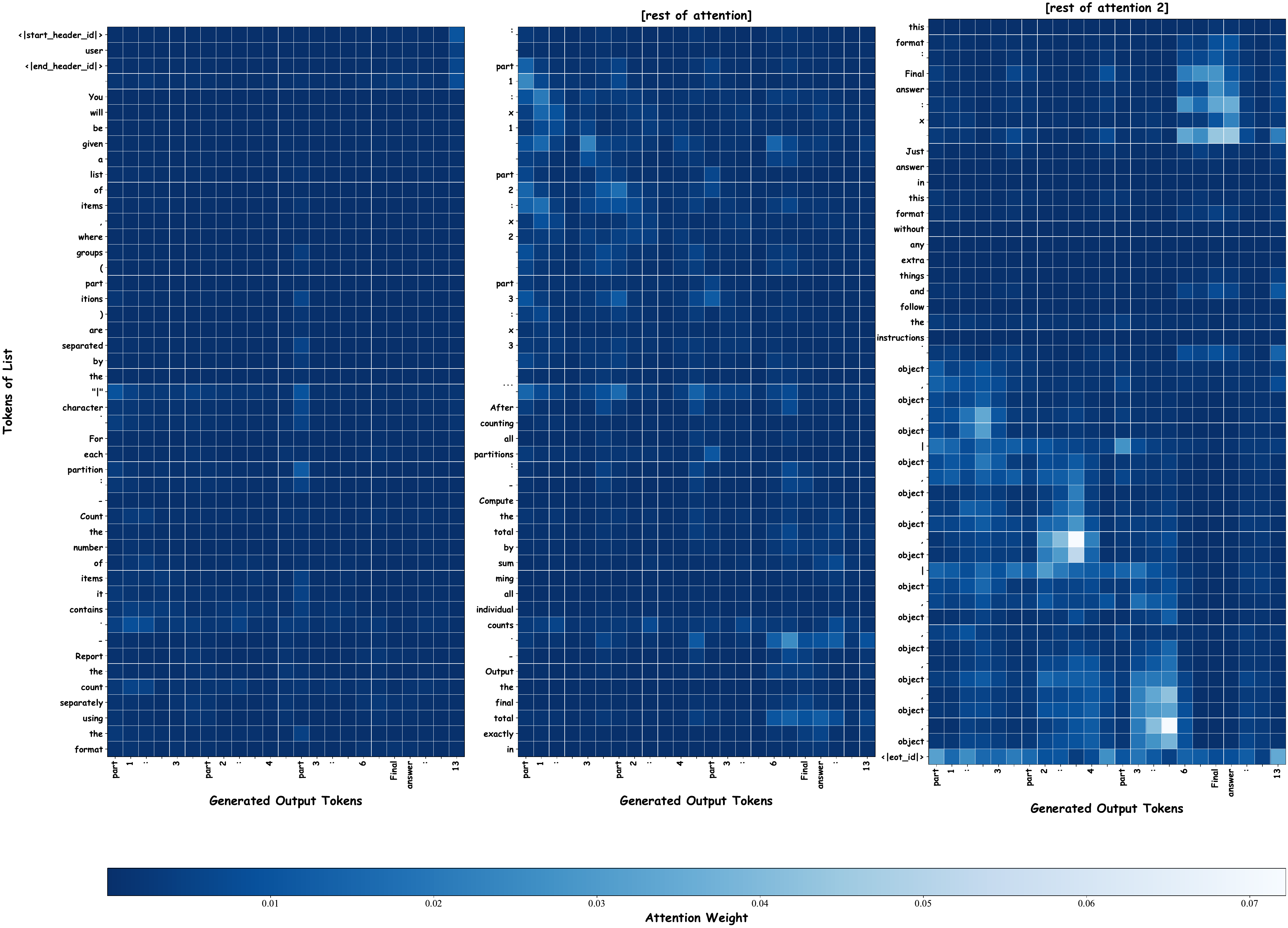}
    \caption{Full attention map for Llama3.2-8B. The x-axis corresponds to output tokens and the y-axis to input tokens. The token immediately preceding the generated number of parts exhibits the strongest attention toward the final item and trailing comma of each partition matches its own segment.}
    \label{fig:llama_full}
\end{figure*}

\begin{figure*}[t]
    \centering
    \includegraphics[width=0.9\linewidth]{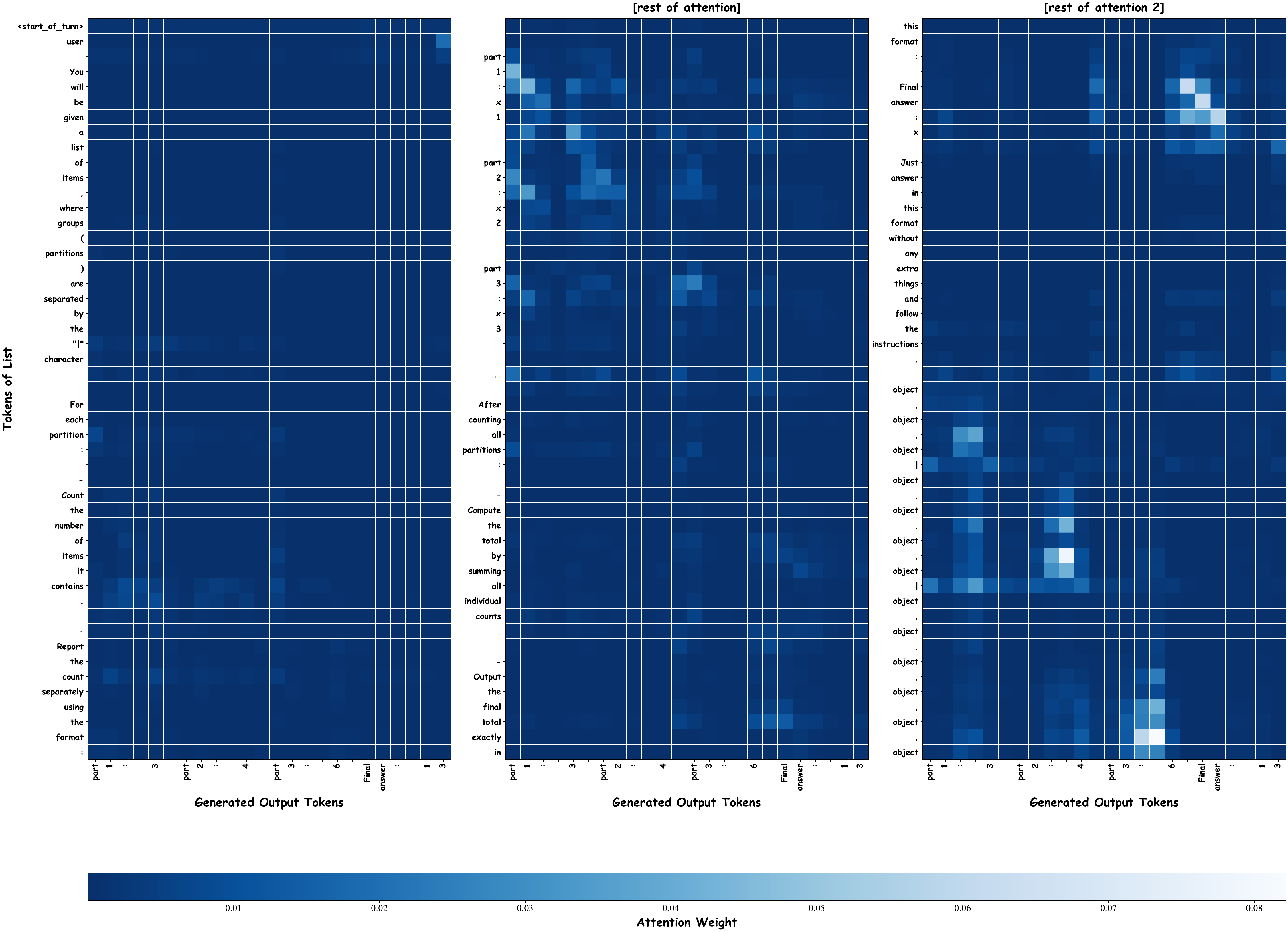}
    \caption{Full attention map for Gemma3-4B. The x-axis corresponds to output tokens and the y-axis to input tokens. The token immediately preceding the generated partition-level count exhibits the strongest attention toward the final item and trailing comma of each partition matches its own segment.}
    \label{fig:gemma_full}
\end{figure*}

Moreover, Figures~\ref{fig:qwen_full}, \ref{fig:llama_full}, and \ref{fig:gemma_full} illustrate the full attention maps for Qwen2.5-7B, Llama3.2 8B, and Gemma3-4B, respectively. In each figure, the x-axis corresponds to output tokens and the y-axis corresponds to input tokens. Across all models, we observe a consistent pattern: the token immediately preceding the generated number of parts exhibits the highest attention weights toward the last item and the final comma of the desired partition. This behavior supports our hypothesis that models rely heavily on attention to the final item–comma structure of each part when determining and generating the number of items for each part.

\subsection{Causal Mediation Analysis}

\paragraph{Token-Level Information Probing}
To decode the latent numbers of specific tokens, we employed CountScope. We modified the original source prompt to facilitate accurate counting by partitioning the input list using vertical bar delimiters ($|$). Additionally, we utilized a monotypic, question-first configuration. The exact prompt structure is provided below:
\begin{lstlisting}
Answer the question with just a number only (We've separated each group of items with "|" so you can calculate the final count easier).
Question: How many fruits are there in the following sentence?
 apple, apple, apple | apple, apple, apple, apple, apple | apple, apple, apple, apple
\end{lstlisting}
To mitigate token-specific artifacts, we performed all experiments using a set of distinct items. (see Appendix~\ref{sec:behavioral_setup})


\paragraph{Information Pathway Localization}
To localize the specific attention heads and layers mediating the flow of count information, we performed attention knockout experiments as outlined in Section~\ref{sec:attention_knockout}. This analysis utilizes a prompt that elicits specific System-2 behavior by explicitly separating partition counting from final aggregation. We employed a strict instructional format forcing the model to output intermediate counts before the final total. This allows us to measure the causal impact of blocking specific attention heads on the accuracy of both intermediate transfer and final summation (the exact prompt template is provided below).
\begin{lstlisting}
I will provide an input text containing fruits separated by the '|' delimiter. Your goal is to count the items in each section and provide a summation.

Input Text:
 apple, apple, apple | apple, apple, apple, apple, apple | apple, apple, apple, apple

Task:
1. Identify each partition separated by '|'.
2. Count the number of fruits in each specific partition.
3. Sum the counts of all partitions.

You must strictly follow this format, adapting the number of parts to the actual input:
part1: <count1>, part2: <count2>, ... , final count: <total_count>
\end{lstlisting}

\end{document}